\documentclass[10pt,twocolumn,letterpaper]{article}

\usepackage[accsupp]{axessibility}
\usepackage{times}
\usepackage{epsfig}
\usepackage{graphicx}
\usepackage{amsmath}
\usepackage{amssymb}
\usepackage{amstext} 
\usepackage{array}   
\newcolumntype{L}{>{$}c<{$}} 
\usepackage{fontawesome5}
\usepackage{siunitx}  
\usepackage{adjustbox}  
\usepackage{tabularx}   
\usepackage{wrapfig}
\usepackage{tikz}
\usepackage{bbding}
\usepackage[utf8]{inputenc}
\usepackage[T1]{fontenc}

\usepackage{algorithm}
\usepackage{algorithmic} 
\usepackage{mathtools}
\usepackage{amsthm}

\usepackage{amssymb} 

\usepackage{subfigure}
\usepackage{booktabs}
\usepackage{witharrows}
\usepackage{rotating}  
\usepackage{multirow}
\usepackage{epsfig}

\usepackage{amssymb}
\usepackage{mathtools}
\usepackage{amsmath,bm}

\usepackage{booktabs}
\usepackage{multirow}
\usepackage{adjustbox}
\usepackage{colortbl}
\definecolor{Gray}{gray}{0.9}
\usepackage{xcolor}
\usepackage{color,colortbl}

\NewDocumentCommand{\qz}{ mO{} }{\textcolor{blue}
{\textsuperscript{\textit{QZ}}\textsf{\textbf{\small[#1]}}}}
\NewDocumentCommand{\xt}{ mO{} }{\textcolor{red}{\textsuperscript{\textit{XT}}\textsf{\textbf{\small[#1]}}}}
\NewDocumentCommand{\zy}{ mO{} }{\textcolor{green}{\textsuperscript{\textit{ZY}}\textsf{\textbf{\small[#1]}}}}
\NewDocumentCommand{\zhouxq}{ mO{} }{\textcolor{blue}{\textsuperscript{\textit{ZhouXQ}}\textsf{\textbf{\small[#1]}}}}
\NewDocumentCommand{\HHB}{ mO{} }{\textcolor{cyan}{\textsuperscript{\textit{HHB}}\textsf{\textbf{\small[#1]}}}}

\usepackage[pagenumbers]{cvpr} 

\usepackage{graphicx}
\usepackage{caption}

\usepackage{makecell}  

\usepackage[pagebackref,breaklinks,colorlinks,citecolor=cvprgreen]{hyperref}
\definecolor{cvprgreen}{rgb}{0,1,0}

\def\paperID{0000} 
\def\confName{ICCV} 
\def\confYear{2026}

\title{Bird-SR: Bidirectional Reward-Guided Diffusion for\\ Real-World Image Super-Resolution}
\author{Zihao Fan$^{1,\dagger}$,~~~~Xin Lu$^{1,\dagger}$,~~~~Yidi Liu$^{1}$,~~~~Jie Huang$^{1}$,~~~~Dong Li$^{1}$,~~~~Xueyang Fu$^{1,}\textsuperscript{\faEnvelope}$,~~~~Baocai Yin$^{2}$\\ 
$^{1}$MoE Key Laboratory of Brain-inspired Intelligent Perception and Cognition,\\
School of Information Science and Technology,   University of Science and Technology of China\\
$^{2}$iFlytek Research, iFlytek Co., Ltd., Hefei, China\\
{~~ \tt\small \{fanzh03, luxion, liuyidi2023, hj0117, dongli6\}@mail.ustc.edu.cn, xyfu@ustc.edu.cn
}
}

\begin{document}

\maketitle

\vspace*{-6mm}
\begin{abstract}
Powered by multimodal text-to-image priors, diffusion-based super-resolution excels at synthesizing intricate details; however, models trained on synthetic low-resolution (LR) and high-resolution (HR) image pairs often degrade when applied to real-world LR images due to significant distribution shifts.
We propose \textbf{Bird-SR}, a \textbf{bi}directional \textbf{r}eward-guided \textbf{d}iffusion framework that formulates super-resolution as trajectory-level preference optimization via reward feedback learning (ReFL), jointly leveraging synthetic LR-HR pairs and real-world LR images.
For structural fidelity easily affected in ReFL, the model is directly optimized on synthetic pairs at early diffusion steps, which also facilitates structure preservation for real-world inputs under smaller distribution gap in structure levels.
For perceptual enhancement, quality-guided rewards are applied to both synthetic and real LR images at the later trajectory phase.
To mitigate reward hacking, the rewards for synthetic results are formulated in a relative advantage space bounded by their ground-truth counterparts, while real-world optimization is regularized via a semantic alignment constraint.
Furthermore, to balance structural and perceptual learning, we introduce a dynamic fidelity-perception weighting strategy that emphasizes structure preservation at early stages and progressively shifts focus toward perceptual optimization at later diffusion steps.
Extensive experiments on real-world SR benchmarks demonstrate that Bird-SR consistently outperforms state-of-the-art methods in perceptual quality while preserving structural consistency, validating its effectiveness for real-world super-resolution. 
Our code can be obtained at \url{https://github.com/fanzh03/Bird-SR}.
\end{abstract}
\vspace*{-6mm}

\section{Introduction}
\label{sec:intro}

\begin{figure}[!t]
  \centering
  \includegraphics[width=\columnwidth]{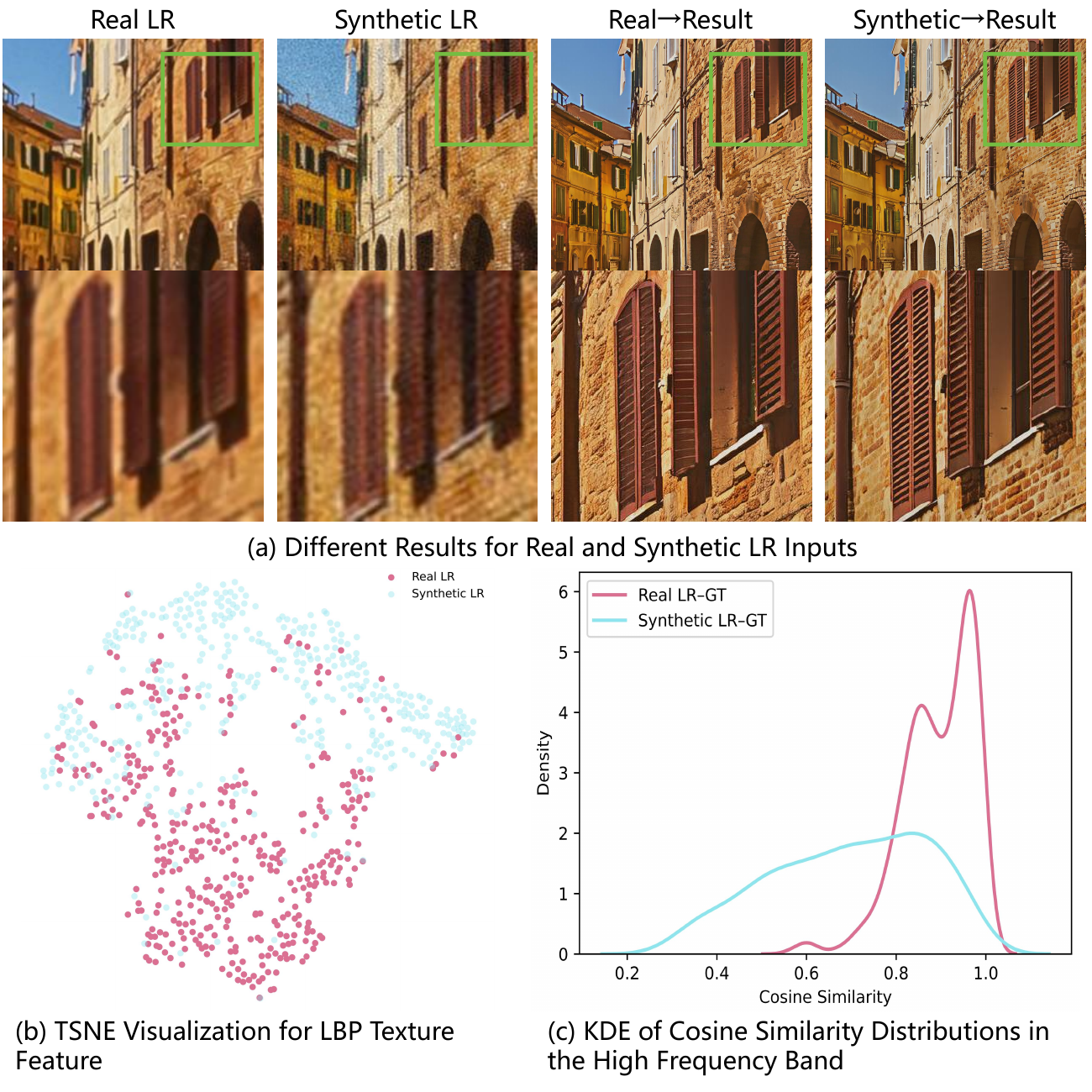}
  \caption{\textit{Motivation.} (a) Due to the degradation gap between synthetically generated low-resolution and real-world low-resolution data, models trained on synthetic datasets tend to produce blurred details when applied to real-world low-resolution images, resulting in an input distribution shift issue. (b) The t-SNE visualization of Local Binary Pattern (LBP) texture features shows clearly separated clusters, indicating a significant texture distribution gap and input distribution shift between synthetic and real-world LR data. (c) Kernel density estimation (KDE) of cosine similarity distributions between low-resolution and high-resolution image pairs in high frequency band. While real-world LR images exhibit a sharp, high-similarity alignment with the ground truth (GT) characterized by a distinct multimodal distribution, synthetic LR images generated by conventional degradation models show a significant distribution broadening and a lower similarity mean. This discrepancy highlights the severe domain gap and input distribution shift that hinders the generalization of models trained solely on synthetic data.}
  \label{fig:analysis}
\end{figure}

In recent years, Diffusion Models~\cite{DDPM,song2021scorebased} have achieved remarkable progress in image generation by learning complex data distributions through progressive denoising. Building on the success of large-scale text-to-image diffusion models~\cite{sd,sd3,sdxl,flux,fang2024vividvideovirtualtryon}, their strong generative priors have been increasingly adopted in image restoration tasks. In diffusion based super-resolution, the LR image serves as a conditioning signal and the reconstruction is formulated as conditional diffusion sampling, enabling pretrained diffusion models to synthesize more natural, diverse, and detailed textures than conventional supervised methods. Despite these advances, existing approaches rely heavily on synthetic paired datasets, and when applied to real world LR images, they often suffer from distribution shift, leading to the loss of high-frequency fine details (as visualized in Figure~\ref{fig:analysis}).

Meanwhile, recent advances in reward-based preference optimization provide a mechanism for directly optimizing generative models using reward signals, enabling diffusion models to align with human perceptual preferences. 
Despite these advantages, directly applying preference optimization to diffusion models remains highly non-trivial.
\begin{figure}[!t]
  \centering
  \includegraphics[width=\columnwidth]{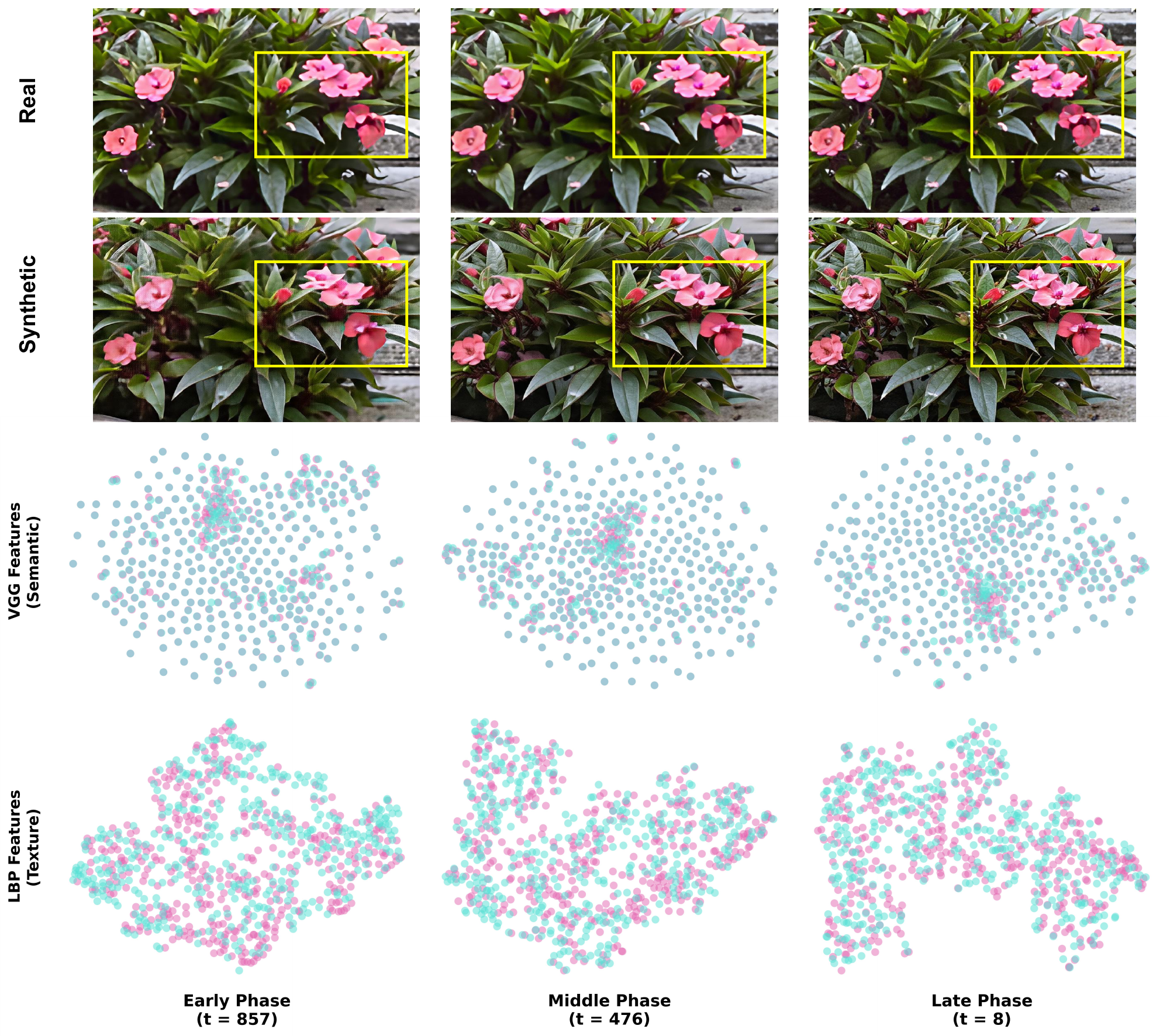}
  \caption{\textit{Evolution of semantic and texture feature spaces during the reverse diffusion process.} We visualize the t-SNE of intermediate predictions ($\hat{x}_0$) from real (red) and synthetic (cyan) reverse trajectories across early, middle, and late denoising stages. Top: VGG features demonstrate that macroscopic semantic structures remain highly consistent throughout the entire process. Bottom: LBP features reveal that, compared to semantic features, texture information presents a distinct domain gap. Moreover, as the reverse timestep progresses, the domains become increasingly dispersed.}
  \label{fig:tsneanalysis}
\end{figure}
Current diffusion based SR and reward optimization frameworks face three critical obstacles: 
(1) \textit{Distribution Shift Induced by Synthetic Paired Data}: Training is predominantly conducted on synthetically generated LR-HR pairs, whereas real world LR degradations are far more diverse and difficult to model, as illustrated in Figure~\ref{fig:analysis}. This mismatch leads to substantial input distribution shifts, causing models to hallucinate over-smoothed or unrealistic textures. 
(2) \textit{Reward Hacking under Perceptual Optimization}: In real-world images, the lack of paired supervision makes the model particularly prone to over-optimizing the reward signal, leading to visually implausible artifacts. 
(3) \textit{Trajectory-Level Trade-off in Diffusion}: Finally, since different stages of the diffusion trajectory contribute differently to structure and appearance, blindly enforcing reward optimization across all steps may amplify perceptual gains at the cost of structural fidelity.

To address these challenges, we propose a bidirectional reward-guided diffusion super-resolution framework that jointly optimizes the forward and reverse processes over both synthetic and real-world data, enabling effective alignment of real-world SR capability, as illustrated in Figure~\ref{fig:pipeline}. Our method is built upon three core ideas: 
(1) Direct optimization via noise injection and relative reward for synthetic data. We inject noise into clean images and directly recover them with a closed-form single-step forward formula. Leveraging the available ground truth, we enable precise supervision of structural distortion and optimize the model using relative rewards. This significantly improves optimization stability, enables flexible timestep control, and further mitigates reward hacking for real-world data.
(2) Reward feedback learning for real-world low-resolution data with semantic alignment. Figure~\ref{fig:tsneanalysis} shows that during the reverse denoising process, while the semantic representations of synthetic and real data are highly aligned, their high-frequency perceptual details manifest a pronounced domain divergence. 
Motivated by this observation, we perform reward feedback along the reverse denoising trajectory, applying the reward signal at the later steps to optimize perceptual detail generation. This naturally corresponds to the temporal mechanism of diffusion models, which generate semantics in the early stages and synthesize textures in the later stages. Meanwhile, we incorporate semantic supervision to constrain structural consistency.
(3) Trajectory-aware distortion-perception weighting along the forward diffusion trajectory. We introduce dynamic weighting to balance structural learning at early steps and perceptual learning at later steps for synthetic data.

Our contributions can be summarized as follows:
\begin{itemize}
\item We propose a bidirectional, reward-guided diffusion super-resolution framework that jointly optimizes synthetic and real-world data via coordinated forward and reverse processes, effectively aligning super-resolution behavior under real-world input distribution shifts.
\item For synthetic data, we introduce a stable reward optimization strategy based on noise injection and relative rewards along the forward diffusion trajectory, by incorporating a trajectory-aware dynamic fidelity-perception weighting.
\item For real-world data, we propose reverse trajectory reward feedback learning with semantic alignment, applying reward signals to refine perceptual details, while incorporating semantic supervision to preserve structural consistency.
\item Extensive experimental results show that our method delivers more realistic and detailed super-resolution results on real world images.
\end{itemize}

\section{Related Work}
\label{sec:related}
\subsection{Diffusion based Super-Resolution}
Early CNN-based~\cite{SRCNN,VDSR,EDSR,RCAN,8954252}  and transformer-based~\cite{10.1007/978-3-031-19790-1_39,IPT,chen2023activating,10824429,9607618} image super-resolution (ISR) methods focus on optimizing pixel-wise fidelity (e.g., PSNR/SSIM) via direct LR to HR mapping, but often yield overly smooth results under complex and unknown real-world degradations.
To handle real-world inputs, Real-ISR approaches simulate diverse degradation processes during training, with representative methods including BSRGAN~\cite{BSRGAN} and Real-ESRGAN~\cite{realesrgan}, while GAN-based methods~\cite{GLEAN,HGGT,10.1007/978-3-031-19797-0_33} such as ESRGAN~\cite{esrgan} further enhance the perceptual quality through adversarial learning.

Recent diffusion-based ISR methods~\cite{DoSSR,TSD-SR,SinSR,AdcSR,OSEDiff,FaithDiff,cheng2025effective,ControlSR,sun2023ccsr} leverage the strong generative priors of pre-trained text-to-image (T2I) diffusion models~\cite{sd,sd3,sdxl,flux} to address Real-ISR. 
StableSR~\cite{StableSR} injects LR structural information into Stable Diffusion via ControlNet~\cite{ControlNet}, while DiffBIR~\cite{DiffBIR} and PASD~\cite{PASD} adopt degradation-restoration pipelines followed by conditional diffusion-based detail enhancement. 
Beyond pixel-level conditioning, SeeSR~\cite{SeeSR}, CoSeR~\cite{CoSeR}, and PiSA-SR~\cite{PiSA-SR} introduce language and vision collaboration to enable semantic-aware and controllable diffusion.
From an efficiency perspective, ResShift~\cite{ResShift} embeds LR information into the initial noise space, whereas InvSR~\cite{InvSR} and UPSR~\cite{UPSR} further refine noise prediction and inversion strategies.
Scaling model and data size, as demonstrated by SUPIR~\cite{SUPIR} and DreamClear~\cite{DreamClear}, significantly improves photorealism. DiT4SR~\cite{DiT4SR} introduces the Diffusion Transformer~\cite{dit} (DiT) into SR by incorporating an LR-conditioned stream within DiT blocks.

\begin{figure*}[!t]
\setlength{\abovecaptionskip}{0.1cm}
  \begin{center}
    \includegraphics[width=1.0\linewidth]{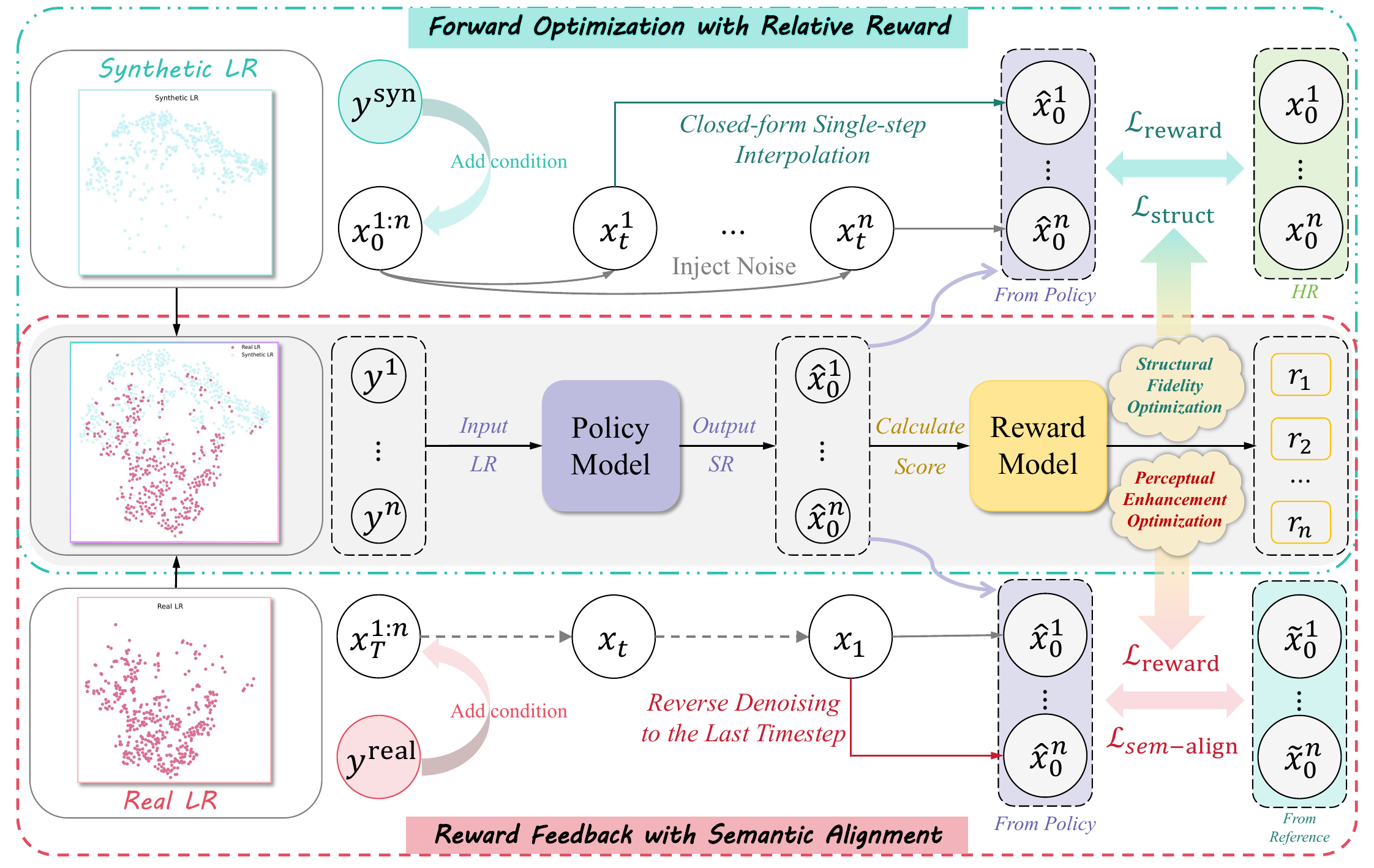}
  \end{center}
  \caption{Overview of the proposed Bird-SR, a bidirectional reward-guided diffusion framework for real-world super-resolution. For synthetic low-resolution data, predefined noise is injected into clean images $x_0^{1:n}$, and intermediate predictions $\hat{x}_0^{1:n}$ are obtained via closed-form single-step interpolation. For real-world low-resolution data, sampling starts from pure noise $x_T^{1:n}$ and optimizes only the final timestep along the reverse diffusion trajectory, with a reference model output $\tilde{x}_0^{1:n}$ introduced to enforce semantic alignment.
  }
  \label{fig:pipeline}
\end{figure*}

\subsection{Preferences Optimization for Diffusion Models}
Recent studies have explored aligning diffusion models with human preferences or task-specific objectives through preference optimization.
One line of work focuses on direct fine-tuning with scalar reward signals, where differentiable rewards are backpropagated through the denoising process.
Representative methods include ReFL~\cite{refl}, DRaFT~\cite{draft}, and AlignProp~\cite{alignprop}, which inject reward feedback into diffusion training to guide perceptual or semantic improvements, while employing truncated backpropagation or memory-efficient techniques to alleviate optimization instability.
Another line of research aligns diffusion models from a reinforcement learning perspective using policy gradient–based methods. DDPO~\cite{ddpo} and DPOK~\cite{dpok} treat the reverse diffusion process as a sequential decision trajectory and optimize model parameters via expected rewards. Diffusion-DPO~\cite{diffusion-dpo} extends direct preference optimization to diffusion models using pairwise preferences, avoiding explicit reward modeling. More recently, GRPO-based strategies~\cite{grpo} improve training stability and scalability, with Flow-GRPO~\cite{flow-grpo} and Dance-GRPO~\cite{dancegrpo} demonstrating effectiveness for large-scale visual generation and preference alignment. SRPO~\cite{srpo} introduces Direct-Align, which predefines a noise prior and formulates rewards as text-conditioned signals.
In contrast to prior approaches, our method adopts a mixed forward–backward optimization framework that enables stable and timestep-aware reward feedback, leading to improved robustness in real-world image super-resolution.

\section{Preliminary}
\label{sec:pre}

\subsection{Conditional Diffusion for Super-Resolution}
Denoising Diffusion Probabilistic Models (DDPMs)~\cite{DDPM,song2021scorebased} are generative models that learn complex data distributions through a progressive denoising process. 
Given a high-resolution image $x_0$, the forward diffusion process gradually adds gaussian noise over $T$ timesteps:
\begin{equation}
x_t = \alpha_t x_0 + \sigma_t \epsilon, \quad \epsilon \sim \mathcal{N}(0, I), \; t = 1, \dots, T,
\label{equ:forward}
\end{equation}
where $\alpha_t$ and $\sigma_t$ control the noise schedule. 
The reverse process aims to reconstruct $x_0$ from $x_T$ by iteratively denoising:
\begin{equation}
p_\theta(x_{t-1}|x_t) = \mathcal{N} (\mu_\theta(x_t), \Sigma_\theta(x_t)),
\end{equation}
where $p_\theta$ is parameterized by a neural network $\epsilon_\theta(x_t,t)$ trained to predict the noise added at each timestep.
Conditional diffusion extends DDPMs to image restoration tasks by conditioning the denoising process on a low-resolution input $y$:
\begin{equation}
p_\theta(x_{t-1}|x_t,y) = \mathcal{N} (\mu_\theta(x_t|y), \Sigma_\theta(x_t|y)).
\label{equ:reverse-form}
\end{equation}
where the low-resolution signal $y$ is incorporated into the denoising network via condition injection, resulting in the conditional model $\epsilon_\theta(x_t, t \mid y)$.

\subsection{Reward Feedback Learning}
To align diffusion models with human preferences or specific perceptual criteria, reward feedback learning fine-tunes the model parameters $\theta$ by maximizing a reward function $r(x_0)$ defined on the generated outputs.
The iterative denoising process from $x_T$ to $x_0$ is treated as a differentiable computation graph, which enables gradients from the reward signal to be backpropagated through the sampling procedure for direct optimization.
However, backpropagating through the full $T$-step sampling trajectory incurs substantial memory overhead and often leads to training instability.
To address this issue, the computation graph is truncated by propagating gradients only to a randomly selected late-stage timestep $t$.
The resulting objective can be formulated as:
\begin{equation}
\mathcal{L} = \lambda \mathbb{E}_{y}\big[\phi\big(r(\hat{x}_0^t(x_t, t))\big)\big],
\end{equation}
where $\phi(\cdot)$ denotes a preference loss function, and $\hat{x}_0^{t}(x_t, t)$ represents the predicted clean image at timestep $t \in [t_{\min}, t_{\max}]$, obtained either by truncating the sampling chain at step $t$ and directly predicting $x_0$, or by completing the full sampling trajectory while stopping gradient propagation beyond timestep $t$.

\section{Method}
In this section, we present \textbf{Bird-SR}, a reward-guided diffusion framework for real-world image super-resolution, as illustrated in Figure~\ref{fig:pipeline}.
The key idea of our approach is to align diffusion-based super-resolution models with the trajectory-level preferences, jointly leveraging
synthetic LR-HR pairs and real-world LR images through stable bidirectional reward feedback.

\subsection{Forward Optimization with Relative Reward}
\label{sec:sys-data-forward}

For synthetic paired data, we adopt a directly injected noise optimization strategy.
For forward processes, instead of performing full diffusion sampling, we inject predefined gaussian noise into clean high-resolution images at a randomly selected timestep.
Given a clean high-resolution image $x_0$ and its corresponding low-resolution image $y$, with the forward diffusion process injecting gaussian noise at timestep $t$ as defined in Eq.~\ref{equ:forward}, the model is initialized using a pre-trained conditional denoising network $\epsilon_\theta(x_t, t \mid y)$.
Instead of performing multi-step reverse diffusion, we reconstruct $x_0$ using a closed-form single-step interpolation:
\begin{equation}
\hat{x}_0 = \frac{x_t-\sigma_t\epsilon_\theta(x_t,t\mid y)}{\alpha_t}.
\label{eq:closed-form}
\end{equation}
This formulation eliminates the need to backpropagate through long diffusion chains, significantly stabilizing reward optimization. Moreover, it enables flexible control over reward feedback at arbitrary timesteps, allowing the model to learn timestep-aware behaviors under reward supervision. 

Based on paired data, we adopt a relative reward formulation, defined as the difference between the reward of the ground-truth $x_0$ and that of the model prediction $\hat{x}_0$, which encourages the model to improve perceptual quality relative to the ground truth and helps prevent reward hacking. The corresponding optimization objective is given by
\begin{equation}
\mathcal{L}_{\text{pair}} = \mathbb{E}_{t,\epsilon}\big[\phi(r(x_0) - r(\hat{x}_0)) \big].
\label{eq:Lpair}
\end{equation}
where $r$ denotes a reward function that measures the perceptual quality of an image, and $\phi$ denotes the preference loss function. In addition, perceptual learning on high-resolution images facilitates optimization for real-world data.

\subsection{Reward Feedback with Semantic Alignment}
\label{sec:real-world-reverse}
To improve the model’s adaptation to the distribution of real-world low-resolution images, we introduce reward feedback learning to fine-tune the parameters $\theta$ of the super-resolution model. Starting from $x_T$ and proceeding through the reverse diffusion process ($x_T \rightarrow x_{T-1} \rightarrow \cdots \rightarrow x_1$), we optimize the model predictions $x_1$. We define the image at timestep $t=0$ and the loss form is written as:
\begin{equation}
\mathcal{L}_{\text{unpair}} = \mathbb{E}_{t}\big[\phi(r(\hat{x}_0)) \big],
\label{eq:Lunpair}
\end{equation}
where the estimate $\hat{x}_0$ is obtained via Eq.~\eqref{equ:reverse-form}, $r$ denotes a reward function that measures the perceptual quality of an image, and $\phi$ denotes preference loss function. We focus on the last timestep for reward supervision because early timesteps mainly define global structure, which is already well-aligned with real-world data as shown in Figure~\ref{fig:tsneanalysis}, and only the prediction at last timestep is optimized. 

While reward-guided optimization encourages perceptually favorable outputs, only applying reward optimization often leads to reward hacking, especially for real-world LR images with complex and unknown degradation patterns.
To address this issue, we incorporate semantic alignment into the reward feedback learning process. Specifically, for real-world LR images, we leverage DINO-based~\cite{oquab2023dinov2} spatial semantic features to guide the reward guided reverse diffusion process:
\begin{equation}
\mathcal{L}_{\text{sem-align}} = \mathbb{E}_{t}\big[||f(\hat{x}_0)-f(\tilde{x}_0)||^2_2 \big],
\label{eq:Lsem}
\end{equation}
where $f$ denotes the spatial semantic feature extractor, and $\tilde{x}_0$ is generated by the reference model. The resulting semantic representations provide high-level structural constraints, thereby enforcing semantic consistency between the optimized outputs and the reference model predictions, while preserving flexibility for pixel-level perceptual refinement. The final backward objective is defined as
\begin{equation}
\mathcal{L}_{\text{reverse}} = 
\mathcal{L}_{\text{unpair}}(\theta) + \lambda_\text{sem} \mathcal{L}_{\text{sem-align}}(\theta),
\label{eq:Lreverse}
\end{equation}
where $\lambda_\text{sem}$ is a hyperparameter controlling the relative importance of semantic alignment.

\begin{figure*}[!t]
      \begin{center}
    \includegraphics[width=1.0\linewidth]{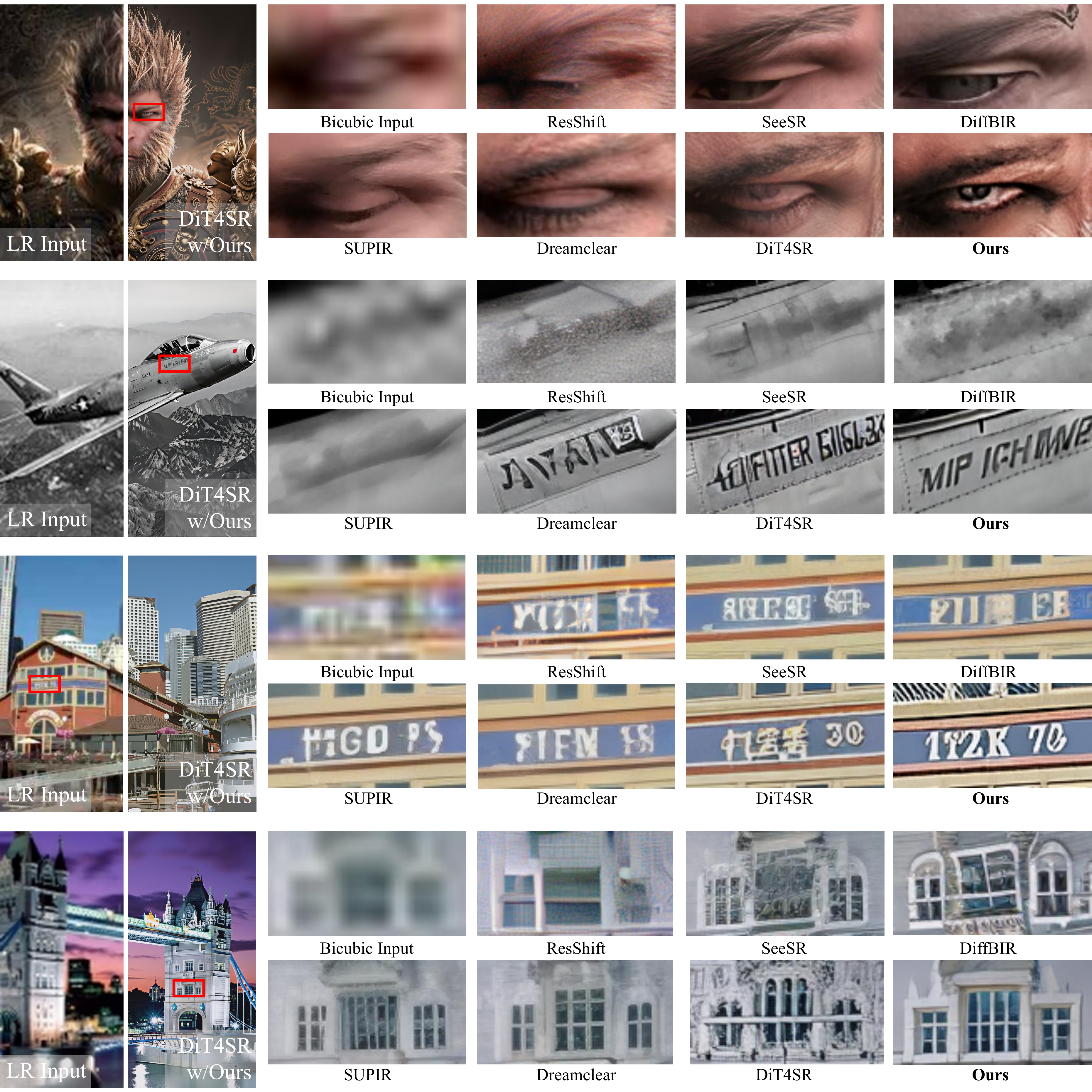}
      \end{center}

    \caption{Qualitative comparisons with state-of-the-art Real-ISR methods. Our method performs best in terms of image realism and detail generation especially preserving fine structures and restoring text details. More visual results are in the \textbf{Appendix}.}
    \label{fig:vision}
    \vspace{-1em}
\end{figure*}

\subsection{Dynamic Distortion-perception Weight}
\label{Weight}
Reward-guided optimization primarily emphasizes perceptual quality, while conventional distortion-based losses (e.g., $\ell_1$ or $\ell_2$) focus on pixel-wise fidelity.
However, the relative importance of distortion and perception varies across diffusion timesteps.
Early timesteps tend to preserve global structures and low-frequency content, where distortion supervision is crucial for stabilizing reconstruction, whereas later timesteps mainly refine high-frequency details, where perceptual rewards play a more dominant role.

To balance these complementary objectives, we introduce a dynamic distortion--perception weighting strategy that adapts across timesteps.
Specifically, we combine the distortion loss and the reward-based objective using a timestep-dependent weight:
\begin{equation}
\mathcal{L}_{\text{struct}}=\mathbb{E}_{t,\epsilon}\big[ D(\hat{x}_0, x_0)\big],
\label{eq: Lstruct}
\end{equation}
\begin{equation}
\mathcal{L}_{\text{forward}}(\theta,t) = \lambda(t)\mathcal{L}_{\text{pair}}(\theta)+(1-\lambda(t))\mathcal{L}_{\text{struct}}(\theta),
\label{eq: forward}
\end{equation}
where $\mathcal{L}_{\text{struct}}$ denotes a distortion loss computed between $\hat{x}_0$ and HR image $x_0$, $D$ represents the distortion function and $\mathcal{L}_{\text{pair}}$ corresponds to the reward-guided objectives defined in the previous sections.
The weight $\lambda(t) \in [0,1]$ is a monotonically decreasing function of timestep $t$, assigning higher emphasis to distortion at early timesteps and gradually shifting focus toward perceptual rewards at later stages.
This dynamic weighting scheme enables stable optimization by preventing excessive perceptual bias during coarse reconstruction, while still allowing reward feedback to guide fine-grained detail enhancement. Our full method is summarized in Algorithm~\ref{alg:bird-sr-pref}. 

\begin{algorithm}[t]
\footnotesize
\caption{Bidirectional reward-guided diffusion for Real-World Super-Resolution}
\label{alg:bird-sr-pref}
\begin{algorithmic}[1]
\addtolength{\itemsep}{0.2ex}
\REQUIRE 
Synthetic paired dataset $\mathcal{D}_{\mathrm{syn}}=\{(x, y^{\mathrm{syn}})\}$, \\
Unpaired real LR dataset $\mathcal{D}_{\mathrm{real}}=\{y^{\mathrm{real}}\}$, \\
Pre-trained conditional diffusion model $\epsilon_{\theta}$, reward $r$, Initialize the reference model $\epsilon_{\text{ref}}\leftarrow\epsilon_{\theta}$, Semantic features function $f$
\ENSURE Adapted model parameters $\theta$

\FOR{each training iteration}
    \STATE Sample $(x, y) \sim \mathcal{D}_{\mathrm{syn}}$, $t \sim \mathcal{U}(0,T)$, $\epsilon \sim \mathcal{N}(0,I)$
    \STATE Noisy input $x_t = \alpha_t x + \sigma_t \epsilon$
    \STATE Model Predict $\hat{x}_0 =(x_t-\sigma_t\epsilon_\theta(x_t,t\mid y^{\mathrm{syn}}))/{\alpha_t}$ \hfill (Eq. \ref{eq:closed-form})
    \STATE Relative reward $A = r(x)-r(\hat{x}_0)$
    \STATE $\mathcal{L}_{\text{pair}} = \mathbb{E}_{t,\epsilon}\big[\phi(r(x) - r(\hat{x}_0)) \big]$ \hfill (Eq. \ref{eq:Lpair})
    \STATE $\mathcal{L}_{\text{struct}}=\mathbb{E}_{t,\epsilon}\big[ D(\hat{x}_0, x)\big]$ \hfill (Eq. \ref{eq: Lstruct})
    \STATE $\mathcal{L}_{\text{forward}}=\lambda(t)\mathcal{L}_{\text{pair}}(\theta)+(1-\lambda(t))\mathcal{L}_{\text{struct}}(\theta)$ \hfill (Eq. \ref{eq: forward})
    \STATE Update $\theta \leftarrow \theta - \eta \nabla_{\theta} \mathcal{L}_{\text{forward}}$ \hfill\COMMENT{\textcolor{gray}{\textit{Gradient update forward}}}
    
    \STATE Sample $y \sim \mathcal{D}_{\mathrm{real}}$, $x \sim  \mathcal{N}(0,I)$
    \FOR{$t=T$ to $t=j+2$}
    \STATE \textbf{no grad}: reverse for $x_{t-1} \leftarrow x_t$ 
    \ENDFOR
    \STATE \textbf{with grad}:  $x_{j} \leftarrow x_{j+1}$  \hfill (Eq. \ref{equ:reverse-form})
    \STATE Last predict $\hat{x}_0 \leftarrow x_{j}$
    \STATE Reference predict from $\epsilon_{\text{ref}}\longrightarrow{\tilde{x}_0}$
    \STATE Absolute reward $r(\hat{x}_0)$
    
    \STATE $\mathcal{L}_{\text{unpair}} = \mathbb{E}_{t}\big[\phi(r(\hat{x}_0)) \big]$ \hfill (Eq. \ref{eq:Lunpair})
    \STATE $\mathcal{L}_{\text{sem-align}} = \mathbb{E}_{t}\big[||f(\hat{x}_0)-f(\tilde{x}_0)||^2_2 \big]$ \hfill (Eq. \ref{eq:Lsem})
    \STATE $\mathcal{L}_{\text{reverse}}=\mathcal{L}_{\text{unpair}}(\theta) + \lambda_\text{sem} \mathcal{L}_{\text{sem-align}}(\theta)$ \hfill (Eq. \ref{eq:Lreverse})
    \STATE Update $\theta \leftarrow \theta - \eta \nabla_{\theta} \mathcal{L}_{\text{reverse}}$ \hfill\COMMENT{\textcolor{gray}{\textit{Gradient update reverse}}}
\ENDFOR
\STATE \textbf{return} $\theta$
\end{algorithmic}
\end{algorithm}
\section{Experiments}
\label{sec:exp}
\begin{table*}[htbp]
\footnotesize
\renewcommand{\arraystretch}{1.3} 
\setlength{\tabcolsep}{5pt} 
\centering
\begin{tabular}{c|c|ccccc|cc|cc}
\toprule
\multirow{2}{*}{\textbf{Datasets}} & \multirow{2}{*}{\textbf{Metrics}} &\multirow{2}{*}{\textbf{StableSR}} & \multirow{2}{*}{\textbf{DiffBIR}} & \multirow{2}{*}{\textbf{SUPIR}} & \multirow{2}{*}{\textbf{SeeSR}} & \multirow{2}{*}{\textbf{DreamClear}} & \multicolumn{2}{c|}{\textbf{ResShift}}  & \multicolumn{2}{c}{\textbf{DiT4SR}} \\
 & & & & & & & \textbf{Baseline} & \cellcolor[HTML]{F0F8FF}\textbf{w/ Ours} & \textbf{Baseline} & \cellcolor[HTML]{F0F8FF}\textbf{w/ Ours} \\
\midrule
& LPIPS $\downarrow$ & \color[HTML]{FF0000}{0.273} & 0.452 & 0.419 & \color[HTML]{2972F4}{0.317} & 0.354 & 0.353 & \cellcolor[HTML]{F0F8FF}0.362{\color[HTML]{FF8C00}\scriptsize$(\uparrow 0.00)$}  & 0.386 & \cellcolor[HTML]{F0F8FF}{0.382}{\color[HTML]{FF8C00}\scriptsize$(\uparrow 0.01)$} \\
& MUSIQ $\uparrow$   & 58.512 & \color[HTML]{FF0000}{65.665} & 59.744 & 58.512 & 44.047 & 52.392 & \cellcolor[HTML]{F0F8FF}55.701{\color[HTML]{FF8C00}\scriptsize$(\uparrow 3.31)$} & 64.202 & \cellcolor[HTML]{F0F8FF}\color[HTML]{2972F4}{ 64.881}{\color[HTML]{FF8C00}\scriptsize$(\uparrow 0.68)$} \\
& MANIQA $\uparrow$  & 0.559 & \color[HTML]{2972F4}{0.629} & 0.552 & 0.605 & 0.455 & 0.476 & \cellcolor[HTML]{F0F8FF}0.501{\color[HTML]{FF8C00}\scriptsize$(\uparrow 0.03)$} & 0.622 & \cellcolor[HTML]{F0F8FF}\color[HTML]{FF0000}{ 0.630}{\color[HTML]{FF8C00}\scriptsize$(\uparrow 0.01)$} \\
& ClipIQA $\uparrow$ & 0.438 & \color[HTML]{FF0000}{0.572} & 0.518 & 0.543 & 0.379 & 0.379 & \cellcolor[HTML]{F0F8FF}0.422{\color[HTML]{FF8C00}\scriptsize$(\uparrow 0.04)$} & 0.548 & \cellcolor[HTML]{F0F8FF}\color[HTML]{2972F4}{ 0.554}{\color[HTML]{FF8C00}\scriptsize$(\uparrow 0.01)$} \\
    \multirow{-5}{*}{\textbf{DrealSR}}   & LIQE $\uparrow$    & 3.243 & 3.894 & 3.728 & \color[HTML]{FF0000}{4.126} & 2.401 & 2.798 & \cellcolor[HTML]{F0F8FF}3.170{\color[HTML]{FF8C00}\scriptsize$(\uparrow 0.37)$} & 3.881 & \cellcolor[HTML]{F0F8FF}\color[HTML]{2972F4}{ 4.051}{\color[HTML]{FF8C00}\scriptsize$(\uparrow 0.17)$} \\
    \midrule
                                & LPIPS $\downarrow$   & \color[HTML]{2972F4}{0.306 }& 0.347 & 0.357 & \color[HTML]{FF0000}{0.299} & 0.325 & 0.316 & \cellcolor[HTML]{F0F8FF}0.314{\color[HTML]{FF8C00}\scriptsize$(\uparrow 0.01)$} & 0.331 & \cellcolor[HTML]{F0F8FF}{0.322}{\color[HTML]{FF8C00}\scriptsize$(\uparrow 0.02)$} \\
                                & MUSIQ $\uparrow$   & 65.653 & \color[HTML]{2972F4}{68.340} & 61.929 & \color[HTML]{FF0000}{69.675} & 59.396 & 56.892 & \cellcolor[HTML]{F0F8FF}62.171{\color[HTML]{FF8C00}\scriptsize$(\uparrow 5.28)$}  & 66.596  & \cellcolor[HTML]{F0F8FF} {67.257}{\color[HTML]{FF8C00}\scriptsize$(\uparrow 0.66)$}\\
                                & MANIQA $\uparrow$  & 0.622 & 0.653 & 0.574 & 0.643 & 0.546 & 0.511 & \cellcolor[HTML]{F0F8FF}0.561{\color[HTML]{FF8C00}\scriptsize$(\uparrow 0.05)$} & \color[HTML]{2972F4}{0.653} & \cellcolor[HTML]{F0F8FF} \color[HTML]{FF0000}{0.667}{\color[HTML]{FF8C00}\scriptsize$(\uparrow 0.01)$}\\
                                & ClipIQA $\uparrow$ & 0.472 & \color[HTML]{FF0000}{0.586} & 0.543 & \color[HTML]{2972F4}{0.577} & 0.474 & 0.407 & \cellcolor[HTML]{F0F8FF}0.465{\color[HTML]{FF8C00}\scriptsize$(\uparrow 0.06)$}  & 0.550 & \cellcolor[HTML]{F0F8FF}{0.565}{\color[HTML]{FF8C00}\scriptsize$(\uparrow 0.02)$} \\
    \multirow{-5}{*}{\textbf{RealSR}}    & LIQE $\uparrow$    & 3.750 & \color[HTML]{2972F4}{4.026} & 3.780 & \color[HTML]{FF0000}{4.123} & 3.221 & 2.853 & \cellcolor[HTML]{F0F8FF}3.413{\color[HTML]{FF8C00}\scriptsize$(\uparrow 0.56)$} & 3.816 & \cellcolor[HTML]{F0F8FF} {3.944}{\color[HTML]{FF8C00}\scriptsize$(\uparrow 0.13)$} \\
    \midrule
                                & MUSIQ $\uparrow$   & 63.433 & 68.027 & 64.837 & 69.428 & 65.926 & 59.695 & \cellcolor[HTML]{F0F8FF}62.110{\color[HTML]{FF8C00}\scriptsize$(\uparrow 2.42)$} & \color[HTML]{2972F4}{70.180} & \cellcolor[HTML]{F0F8FF}\color[HTML]{FF0000}{70.377}{\color[HTML]{FF8C00}\scriptsize$(\uparrow 0.20)$} \\
                                & MANIQA $\uparrow$  & 0.579 & 0.629 & 0.600 & 0.612 & 0.597 & 0.525 & \cellcolor[HTML]{F0F8FF}0.548{\color[HTML]{FF8C00}\scriptsize$(\uparrow 0.02)$} & \color[HTML]{2972F4}{0.644} & \cellcolor[HTML]{F0F8FF}\color[HTML]{FF0000}{0.652}{\color[HTML]{FF8C00}\scriptsize$(\uparrow 0.01)$} \\
                                & ClipIQA $\uparrow$ & 0.458 & 0.582 & 0.524 & 0.566 & 0.546 & 0.452 & \cellcolor[HTML]{F0F8FF}0.501{\color[HTML]{FF8C00}\scriptsize$(\uparrow 0.05)$} & \color[HTML]{2972F4}{0.588} & \cellcolor[HTML]{F0F8FF}\color[HTML]{FF0000}{0.591}{\color[HTML]{FF8C00}\scriptsize$(\uparrow 0.01)$} \\
    \multirow{-4}{*}{\textbf{RealLR200}} & LIQE $\uparrow$    & 3.379 & 4.003 & 3.626 & 4.006 & 3.775 & 3.054 & \cellcolor[HTML]{F0F8FF}3.350{\color[HTML]{FF8C00}\scriptsize$(\uparrow 0.30)$} & \color[HTML]{2972F4}{4.283} & \cellcolor[HTML]{F0F8FF}\color[HTML]{FF0000}{4.294}{\color[HTML]{FF8C00}\scriptsize$(\uparrow 0.01)$} \\
    \midrule
                                & MUSIQ $\uparrow$   & 56.858 & 69.876 & 66.016 & 70.556 & 66.693 & 59.337 & \cellcolor[HTML]{F0F8FF}64.399{\color[HTML]{FF8C00}\scriptsize$(\uparrow 5.06)$} &  \color[HTML]{2972F4}{71.632} & \cellcolor[HTML]{F0F8FF}\color[HTML]{FF0000}{72.214}{\color[HTML]{FF8C00}\scriptsize$(\uparrow 0.58)$} \\
                                & MANIQA $\uparrow$  & 0.504 & 0.624 & 0.584 & 0.594 & 0.585 & 0.500 & \cellcolor[HTML]{F0F8FF}0.533{\color[HTML]{FF8C00}\scriptsize$(\uparrow 0.03)$} & \color[HTML]{2972F4}{0.632} & \cellcolor[HTML]{F0F8FF}\color[HTML]{FF0000}{0.637}{\color[HTML]{FF8C00}\scriptsize$(\uparrow 0.01)$} \\
                                & ClipIQA $\uparrow$ & 0.382 & 0.578 & 0.483 & 0.562 & 0.502 & 0.417 & \cellcolor[HTML]{F0F8FF}0.488{\color[HTML]{FF8C00}\scriptsize$(\uparrow 0.07)$} & \color[HTML]{2972F4}{0.573} & \cellcolor[HTML]{F0F8FF}\color[HTML]{FF0000}{0.583}{\color[HTML]{FF8C00}\scriptsize$(\uparrow 0.01)$} \\
    \multirow{-4}{*}{\textbf{RealLQ250}} & LIQE $\uparrow$    & 2.719 & 4.003 & 3.605 & 4.005 & 3.688 & 2.753 & \cellcolor[HTML]{F0F8FF}{3.398}{\color[HTML]{FF8C00}\scriptsize$(\uparrow 0.65)$} & \color[HTML]{2972F4}{4.321} & \cellcolor[HTML]{F0F8FF}\color[HTML]{FF0000}{4.399}{\color[HTML]{FF8C00}\scriptsize$(\uparrow 0.08)$} \\
\bottomrule
\end{tabular}
\caption{Quantitative results of Real-ISR methods on four real-world benchmarks based on our method. Best and second best results are highlighted in {\color[HTML]{FF0000} red} and {\color[HTML]{2972F4}blue}, respectively. w/Ours achieves the best or comparable performance across four benchmarks. We supplement our evaluation with two representative diffusion baselines. To account for ResShift’s lower base performance, we compare relative deltas rather than absolute values. {\color[HTML]{FF8C00}Orange} subscripts indicate improvement $(\uparrow \Delta)$ over baselines (2 d.p.). 
}
\label{tab:results-main}
\vspace{-1em}
\end{table*}

\subsection{Experimental Settings}
\textbf{Datasets.} 
We adopt a combination of images from DIV2K~\cite{div2k}, DIV8K~\cite{div8k}, Flickr2K~\cite{flickr2k}, and the first 10K face images from FFHQ~\cite{ffhq} during training. Flickr8K~\cite{Flickr8K} is used for real-world low-resolution dataset. The degradation pipeline of RealESRGAN~\cite{realesrgan} is utilized to synthesize LR-HR training pairs with the same parameter configuration as baseline methods. Following DiT4SR~\cite{DiT4SR}, the resolutions are set to 128 × 128 and 512 × 512 for LR and HR images, respectively. Following ResShift~\cite{ResShift}, the resolutions are set to 64 × 64 and 256 × 256 for LR and HR images. 

Since our method specifically focuses on Real-ISR task, we evaluate our model on four widely used real-world datasets, including DrealSR~\cite{drealsr}, RealSR~\cite{realsr}, RealLR200~\cite{SeeSR}, and RealLQ250~\cite{DreamClear}. All experiments are conducted with the scaling factor of ×4. DrealSR and RealSR respectively consist of 93 and 100 images. Following DiT4SR, center-cropping is adopted for these two datasets, and the resolution of LR images is set to 128 × 128.

\textbf{Metrics.}
Following~\cite{DiT4SR,DreamClear,SUPIR}, PSNR and SSIM~\cite{ssim} are inadequate for measuring perceptual differences. Thus, most studies use the perceptual measurement LPIPS~\cite{lpips} for image fidelity. We use MUSIQ~\cite{musiq}, 
MANIQA~\cite{maniqa}, ClipIQA~\cite{ClipIQA}, and LIQE~\cite{liqe} as non-reference metrics to measure image quality. In addition, we conduct a user study to comprehensively assess both fidelity and perceptual quality.
  
We further validate our method on DiT4SR and ResShift, the \textbf{implementation} details are provided in the \textbf{Appendix}.

\subsection{Comparison with Other Methods}
We compare our method with state-of-the-art RealISR methods, including diffusion-based approaches (i.e., ResShift~\cite{ResShift}, StableSR~\cite{StableSR}, SeeSR~\cite{SeeSR}, DiffBIR~\cite{DiffBIR}, and SUPIR~\cite{SUPIR}), as well as diffusion-based methods with DiT architecture (i.e. DreamClear~\cite{DreamClear} and DiT4SR~\cite{DiT4SR}).
As shown in Table~\ref{tab:results-main}, our method achieves strong quantitative performance across all four real-world super-resolution benchmarks. Additional visual results, including the outputs of ResShift, are presented in the \textbf{Appendix}.
As illustrated in Figure~\ref{fig:vision}, our approach produces more realistic and natural reconstructions in real-world scenarios, with clearer and more faithful details while effectively suppressing artifacts.
These results indicate that our method effectively mitigates the impact of input distribution shift and improves perceptual quality on real-world low-resolution images. 
By leveraging bidirectional reward feedback learning and jointly training on synthetic and real low-resolution data, our method demonstrates superior performance and robustness, all without adding any computational cost or latency during inference.

\subsection{Ablation Experiments}
\label{subsec:ablation}
To further demonstrate the effectiveness of each component, we conduct the ablation study on RealSR~\cite{realsr} with MUSIQ and LPIPS as evaluation metrics. All variants are trained using the same settings as the full model for fair comparison. Additional details are provided in the \textbf{Appendix}.
\begin{table}[!htbp]
    \renewcommand{\arraystretch}{1.1}
    \setlength{\tabcolsep}{5.5pt}
    \centering
    \footnotesize
    \begin{tabular}{c|c|c|c|c}
    \toprule
    Ours vs.       & DIT4SR     & SeeSR     & DiffBIR     & DreamClear    \\
    \midrule
    Realism       & 95.2\% & 98.1\%  & 97.1\%  & 96.2\%   \\
    Fidelity      & 75.2\% & 85.7\%  & 84.8\%  & 82.9\%   \\
    \bottomrule
    \end{tabular}
    \caption{User study results on real-world datasets. The percentages denote the frequency with which Ours was preferred over each compared approach (realism and fidelity).}
    \label{tab:user_study}
\end{table}
\subsubsection{Effectiveness of Bidirectional Reward-Guided Diffusion}
\label{subsubsec:ablation-4}
We first evaluate the contribution of forward and backward processes individually as well as their combination.
The experimental variants are defined as follows:\\
\textbf{1. Only paired forward:} Optimizing the model using only paired synthetic data with forward noise injection.\\
\textbf{2. Only real-world LR reverse:} Optimizing the model using only real-world LR images with reverse process.\\
\textbf{3. All reverse:} Performing reward-guided reverse optimization for both synthetic LR data and real-world LR data.\\
\textbf{4. Mixed forward and reverse:} The full bidirectional optimization framework that jointly leverages synthetic paired data and real-world low-resolution data.

\begin{figure}[tbp]
  \centering
  \includegraphics[width=\columnwidth]{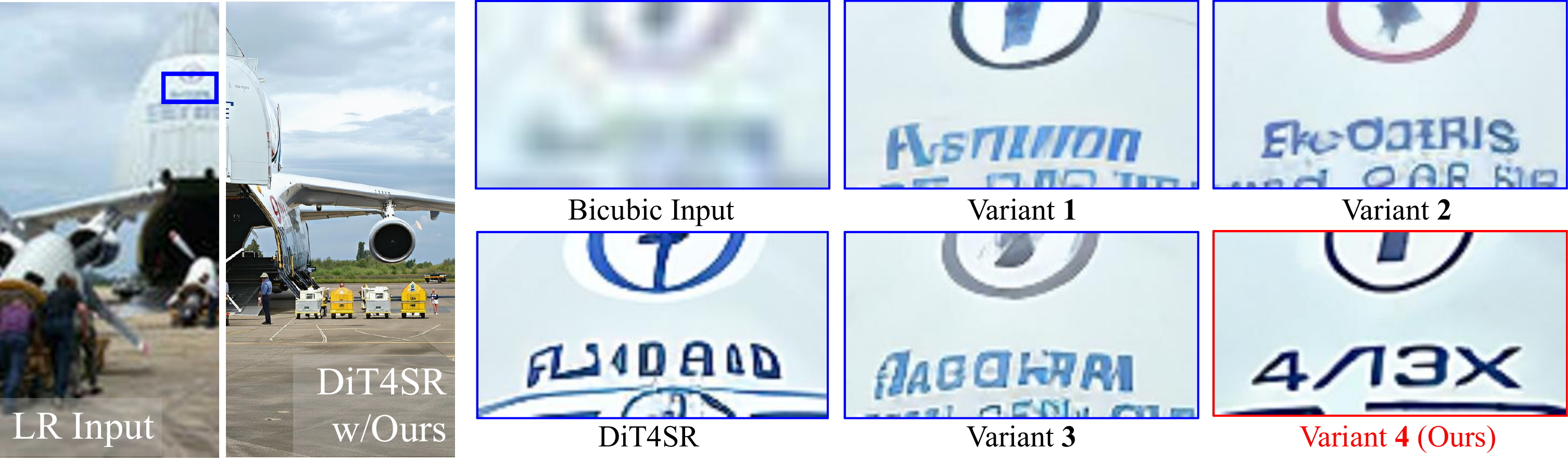}
  \vspace{-1em}
  \caption{Visualization of ablation for the four variants }
  \label{fig:vision_ablation}
\end{figure}

We evaluate these 4 variants in terms of both training cost and performance (MUSIQ / LPIPS scores). 
The results are summarized in Table~\ref{tab:ablation}.
In addition, the visualization of four variants are shown in Figure~\ref{fig:vision_ablation}.

\textbf{Analysis.}
(1) \emph{Only forward} achieves faster training but lacks adaptation to real-world LR degradations, resulting in inferior perceptual quality.
(2) \emph{Only reverse} adapts to real-world LR images but suffers from unstable optimization and an increased risk of reward hacking due to long reverse diffusion trajectories.
(3) \emph{All reverse} improves performance over \emph{Only reverse} but incurs significantly higher computational cost.
(4) \emph{Mixed forward and reverse} achieves the best trade-off between efficiency and perceptual quality, demonstrating the effectiveness of our bidirectional optimization strategy.
These observations indicate that jointly combining forward and reverse optimization not only stabilizes training but also enables better adaptation to real-world LR images.

\begin{table}[ht]
    \footnotesize
    \renewcommand{\arraystretch}{1.1}
    \setlength{\tabcolsep}{15pt}
    \centering
    \begin{tabular}{c|cc|c}
    \toprule
    Setting & MUSIQ $\uparrow$ & LPIPS $\downarrow$ & \begin{tabular}[c]{@{}c@{}}Train cost\end{tabular} \\  \midrule
    1 & 66.687 & 0.344 & 22\% \\
    2 & 67.097 & 0.347 & 100\% \\
    3 & 67.125 & 0.328 & 100\% \\
    4 & \textbf{67.257} & \textbf{0.322} & 64\% \\ \bottomrule
    \end{tabular}
    \caption{Ablation study on RealSR for forward and backward optimization (see sec.\ref{subsubsec:ablation-4}). We report training cost (GPU hours) and perceptual quality measured by MUSIQ and LPIPS. Note that GPU hours are normalized relative to 3 (set to 100\%).     }
    \label{tab:ablation}
\end{table}
\subsubsection{Dynamic Weights.}
\label{subsec:dynamic_weight}
We investigate the impact of different distortion-perception weighting strategies across diffusion timesteps on model performance.
Specifically, we compare several weighting schedules for combining distortion loss $\mathcal{L}_{\text{struct}}$ and reward-based loss $\mathcal{L}_{\text{reward}}$ in our bidirectional reward-guided SR framework:
$\lambda(t) = \left(\frac{t}{T}\right)^\gamma$, which gradually shifts from distortion to perceptual reward, where $\gamma>0$ controls the decay rate.

\begin{figure}[htbp]
  \centering
  \includegraphics[width=\columnwidth]{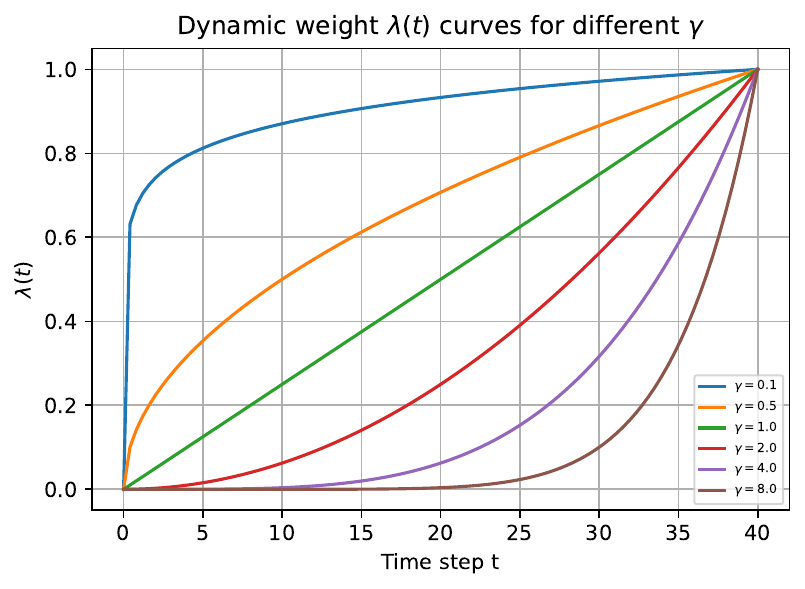}
  \vspace{-2em}
  \caption{Different distortion–perception weighting.}
  \label{fig:ablation_eta_t_curves}
\end{figure}

\textbf{Analysis.}  
Table~\ref{tab:ablation-time} summarizes the quantitative results, and Figure~\ref{fig:ablation_eta_t_curves} visualizes the different weighting schedules. Our ablation study on power-law weighting with varying exponents reveals a critical trade-off~\cite{8578750,RFSR,10.1007/978-3-030-58621-8_37}: stronger structural constraints at early timesteps reduce distortion but limits perceptual reward gains. Conversely, insufficient early-stage constraints lead to structural degradation, which triggers reward hacking. By identifying an optimal weighting scheme, we enhance perceptual preference while preserving structural integrity.  

\begin{table}[htbp]
    \footnotesize
    \renewcommand{\arraystretch}{1.1}
    \setlength{\tabcolsep}{24pt}
    \centering
    
    \begin{tabular}{c|cc}
    \toprule
    Setting & MUSIQ $\uparrow$ & LPIPS $\downarrow$ \\  \midrule
    $\gamma=0.1$ & 67.220 & 0.328 \\
    $\gamma=0.5$ & 67.217 & 0.325 \\
    $\gamma=1.0$ & 67.221 & 0.329 \\
    $\gamma=2.0$ & 67.228 & 0.326 \\
    $\gamma=4.0$ & 67.234 & 0.330 \\
    $\gamma=8.0$ & \textbf{67.257} & \textbf{0.322} \\ \bottomrule
    \end{tabular}

    \caption{Ablation study on different distortion–perception weighting. We report perceptual quality of MUSIQ and LPIPS.}
    \label{tab:ablation-time}
\end{table}

\subsubsection{Ablation on Loss Components.}
To verify the effectiveness of our proposed structural constraint and semantic alignment, we conduct ablation studies on DiT4SR. For structural constraint, we employ perceptual supervision via LPIPS loss between the policy model's output and the ground truth. We deliberately eschew conventional $L_1$ or $L_2$ constraints, as they tend to induce over-smoothed results. For semantic alignment, we leverage DINO in the feature space to align the outputs of the reference and policy models. Since DINO captures high-level semantics, it effectively constrains structural distortions arising from reward hacking while reserving optimization overhead for pixel-level detail preferences. This dual-pronged approach ensures both the stability and efficacy of the training process.

\textbf{Analysis.}  
As presented in Table~\ref{tab:app-ablation-loss-dit}, utilizing solely the reward signal without additional constraints leads to a significant improvement in perceptual metrics. However, this gain is achieved at the expense of severe structural distortion, indicating that the model has succumbed to reward hacking. The introduction of either semantic or structural constraints results in a regression in perceptual scores but successfully mitigates distortion. Ultimately, by integrating both constraints, our model achieves a stable and synergistic improvement across both perceptual quality and distortion metrics.

\begin{table}[htbp]
    \footnotesize
    \renewcommand{\arraystretch}{1.1}
    \setlength{\tabcolsep}{5pt}
    \centering
    \begin{tabular}{c|ccc|cc}
    \toprule
    Model & \begin{tabular}[c]{@{}c@{}} $\mathcal{L}_{\text{struct}}$ \end{tabular} 
          & \begin{tabular}[c]{@{}c@{}} $\mathcal{L}_{\text{sem-align}}$ \end{tabular} 
          & \begin{tabular}[c]{@{}c@{}} $\mathcal{L}_{\text{reward}}$ \end{tabular} 
    & ClipIQA $\uparrow$  & LPIPS $\downarrow$ \\ \midrule
    \rm{A}      & \XSolidBrush   & \XSolidBrush   &  \CheckmarkBold   &0.541   & 0.355  \\
    \rm{B}      & \XSolidBrush   & \CheckmarkBold &  \CheckmarkBold   &0.538   & 0.338  \\ 
    \rm{C}      & \CheckmarkBold & \XSolidBrush   &  \CheckmarkBold   &0.544  & 0.326  \\ \midrule
     FULL       & \CheckmarkBold & \CheckmarkBold &  \CheckmarkBold   &0.565  & 0.322     \\ \bottomrule
    \end{tabular}
    \caption{Ablation results on RealSR for DiT4SR. All
    variants are trained using the same settings as the full model. }
    \label{tab:app-ablation-loss-dit}
\end{table}
\section{Limitations \& Future Work.}
While Bird-SR demonstrates strong performance on real-world super-resolution, its effectiveness is inherently tied to the quality of the reward model. 
The current reward formulation primarily captures global perceptual preferences and may be insufficient for modeling fine-grained, region-specific visual attributes such as local textures, edges, or small structures. 
Future work could explore more precise and fine-grained reward models, for example by incorporating spatially adaptive or multi-scale reward signals, as well as learning region-aware preferences from richer human feedback. 
Such extensions may further enhance the robustness of reward-guided diffusion for real-world image restoration.

\section{Conclusion}
\label{sec:Conclusion}
In this work, we presented Bird-SR, a bidirectional reward-guided diffusion framework for real-world super-resolution that addresses the distribution shift between synthetic training data and real-world low-resolution images. By formulating reward-guided diffusion training as a trajectory-level preference optimization problem, Bird-SR enables stable and efficient alignment of diffusion trajectories with distribution-level perceptual preferences.
The proposed forward and reverse reward-guided processes effectively balance structural fidelity and perceptual realism, while mitigating common issues such as unrealistic artifacts and reward hacking in real-world scenarios. Extensive experiments on real-world super-resolution benchmarks demonstrate that Bird-SR consistently achieves superior perceptual quality without sacrificing structural consistency.
These results suggest that bidirectional reward-guided diffusion provides a promising direction for advancing real-world image restoration beyond conventional supervised paradigms.

\clearpage
\section{Detailed Formulations of Diffusion-based Super-Resolution}
\label{sec:appendix_diffusion_sr}

Diffusion-based super-resolution (SR) methods can be broadly categorized into two classes depending on whether the diffusion model is trained from scratch for the SR task or adapted from a pretrained generative diffusion model. 
In this section, we provide a unified formulation of both paradigms and clarify their key differences.

\subsection{Diffusion Models Trained from Scratch for Super-Resolution}
The first category directly trains a diffusion model for super resolution using paired low-resolution (LR) and high-resolution (HR) data.
Representative methods such as ResShift formulate the forward diffusion process as an interpolation between the HR image $x_0$, the LR-conditioned signal $x_1$, and Gaussian noise.

Specifically, the forward process at timestep $t$ is defined as:
\begin{equation}
x_t = \alpha_t x_0 + \beta_t x_1 + \gamma_t \epsilon, \quad \epsilon \sim \mathcal{N}(0, I),
\end{equation}
where $x_0$ denotes the ground-truth HR image, $x_1$ is an upsampled or encoded representation of the LR input, and $\{\alpha_t, \beta_t, \gamma_t\}$ are predefined or learned coefficients satisfying normalization constraints.

The reverse denoising process aims to gradually reconstruct $x_0$ from $x_T$ conditioned on $y=x_1$, typically by predicting either the noise term $\epsilon$ or the clean image $x_0$:
\begin{equation}
p_\theta(x_{t-1}|x_t,y) = \mathcal{N} (\mu_\theta(x_t|y), \Sigma_\theta(x_t|y)),
\end{equation}
The diffusion model is trained by minimizing a denoising objective over randomly sampled timesteps.
In practice, a commonly used formulation predicts the injected noise $\epsilon$:
\begin{equation}
\mathcal{L}
= \mathbb{E}_{x_0, y, t, \epsilon}
\left[
\left\|
\epsilon - \epsilon_\theta(x_t, t \mid y)
\right\|_2^2
\right],
\end{equation}
where $t$ is uniformly sampled from $\{1, \dots, T\}$.
Alternatively, equivalent objectives can be derived by directly predicting $x_0$ or a hybrid parameterization.

Since the diffusion model is trained end-to-end on paired SR data, this paradigm offers strong task-specific performance but often suffers from limited generalization when applied to real-world LR images with unknown degradations.

\subsection{Pretrained Diffusion Models with Conditional Adaptation}
The second category builds upon large-scale pretrained diffusion models that are originally trained for unconditional or text-to-image generation.
Instead of retraining the diffusion process from scratch, these methods adapt the pretrained model to super-resolution via conditional control mechanisms.

A common approach is to introduce an external conditioning module, such as ControlNet, which injects LR structural information into intermediate layers of the frozen pretrained diffusion backbone.
Alternatively, lightweight parameter-efficient tuning strategies, such as Low-Rank Adaptation~\cite{hu2022lora} (LoRA), are employed to fine-tune a subset of model parameters while preserving the pretrained generative prior.
Specifically, the forward diffusion process remains unchanged and is given by:
\begin{equation}
x_t = \alpha_t x_0 + \sigma_t \epsilon, \quad \epsilon \sim \mathcal{N}(0, I),
\end{equation}
where $x_0$ denotes the target high-resolution image.
The low resolution observation $y$ is not involved in the forward noising process but is provided as a conditioning signal to the denoising network.
Formally, the reverse diffusion process can be written as:
\begin{equation}
p_{\theta,\psi}(x_{t-1}|x_t,y) = \mathcal{N} (\mu_{\theta,\psi}(x_t|y), \Sigma_{\theta,\psi}(x_t|y)).
\end{equation}
where $y$ denotes the LR input, $\theta$ represents the frozen pretrained parameters, and $\psi$ corresponds to the newly introduced or fine-tuned conditional parameters.
Training is performed by optimizing a denoising objective similar to standard DDPMs, while updating only the conditional parameters $\psi$.
In practice, the model is commonly trained to predict the injected noise:
\begin{equation}
\mathcal{L} 
= \mathbb{E}_{x_0, y, t, \epsilon}
\left[
\left\|
\epsilon - \epsilon_{\theta,\psi}(x_t, t \mid y)
\right\|_2^2
\right],
\end{equation}
where $t$ is uniformly sampled from $\{1, \dots, T\}$.
By leveraging the strong prior encoded in large pretrained diffusion models, this paradigm often produces visually appealing results.
However, adapting such models to real-world SR remains challenging due to mismatches between pretrained distributions and real-world degradation patterns, motivating further alignment strategies such as reward-guided optimization.

\section{Implementation details}
\label{sec:app-exp_setting_dataset}

\subsection{Local Binary Pattern (LBP) Texture Features}
\begin{figure*}[!tbp]
  \centering
  \includegraphics[width=1.0\linewidth]{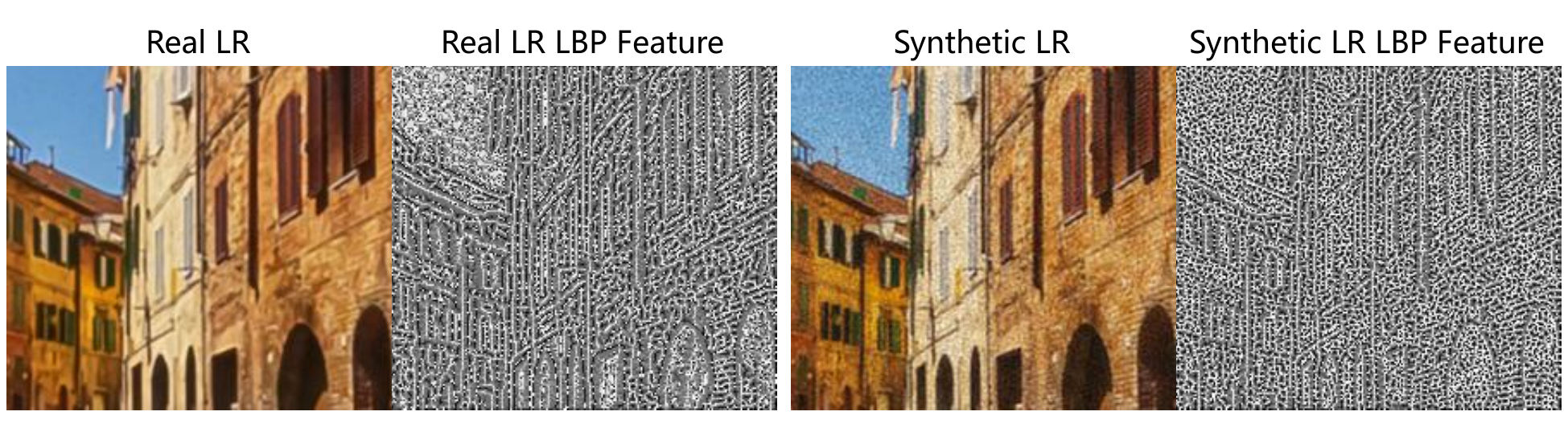}
  \vspace*{-1em}
  \caption{Visualization of LBP Texture Features. As evidenced by the LBP texture results, compared to real-world data, the synthetic LR data is superimposed with additional information in the high-frequency components. This leads to an input distribution shift, particularly hindering the recovery of fine-grained details.}
  \label{fig:add_analysis_LBP}
\end{figure*}

To provide further insights into the texture characteristics of low-resolution and high-resolution images, we visualize the corresponding Local Binary Pattern (LBP) representations as shown in Figure~\ref{fig:add_analysis_LBP}. LBP captures local texture structures by encoding neighborhood intensity variations and is widely used for analyzing fine-grained texture patterns. The visualization highlights the differences in local texture consistency and structural regularity between low-resolution inputs and their high-resolution counterparts, offering an intuitive understanding of texture degradation and restoration behavior.

\begin{figure*}[!t]
  \centering
  \includegraphics[width=1.0\linewidth]{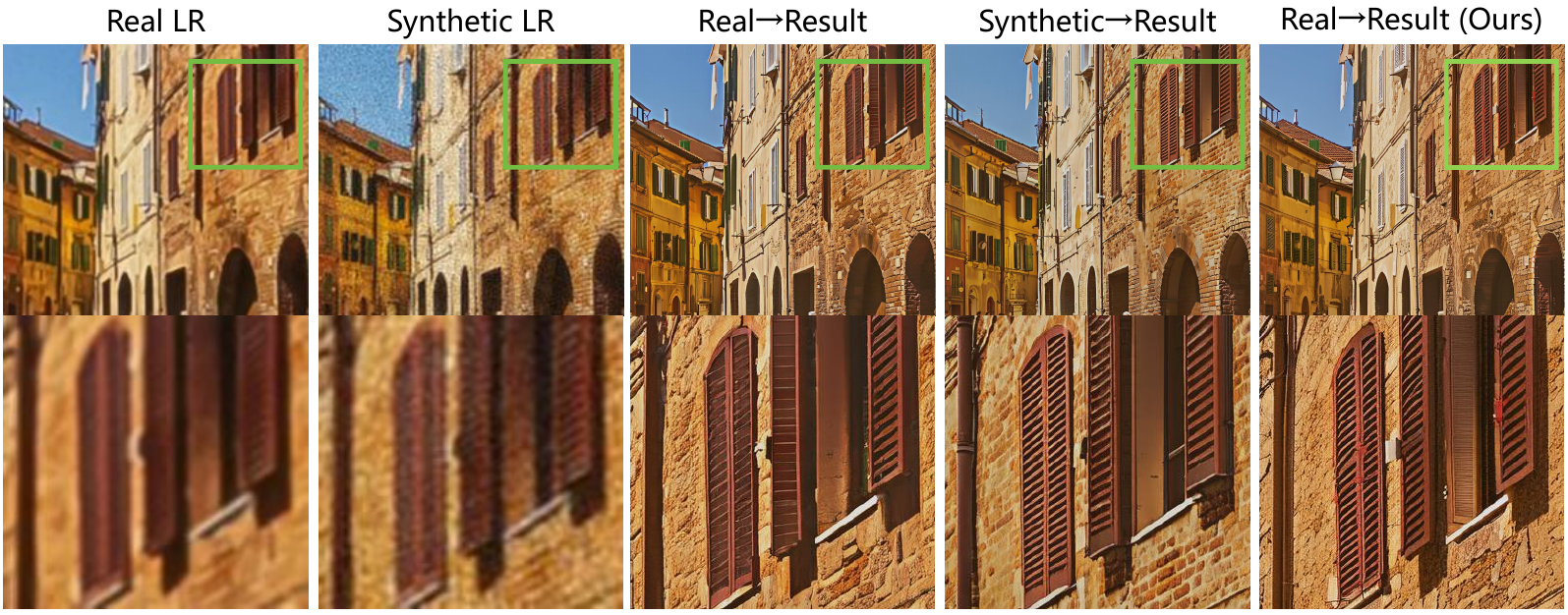}
  \caption{Models trained solely on synthetic data tend to produce blurred details when applied to real-world LR inputs, in contrast to their performance on synthetic LR samples. After fine-tuning with our proposed method, the model adapts effectively to real-world data and is able to recover substantially richer and more faithful details.}
  \label{fig:analysis-app}
\end{figure*}
\subsection{KDE of Cosine Similarity Distributions in Deep Feature Space}
\begin{figure*}[!tbp]
  \centering
  \includegraphics[width=1.0\linewidth]{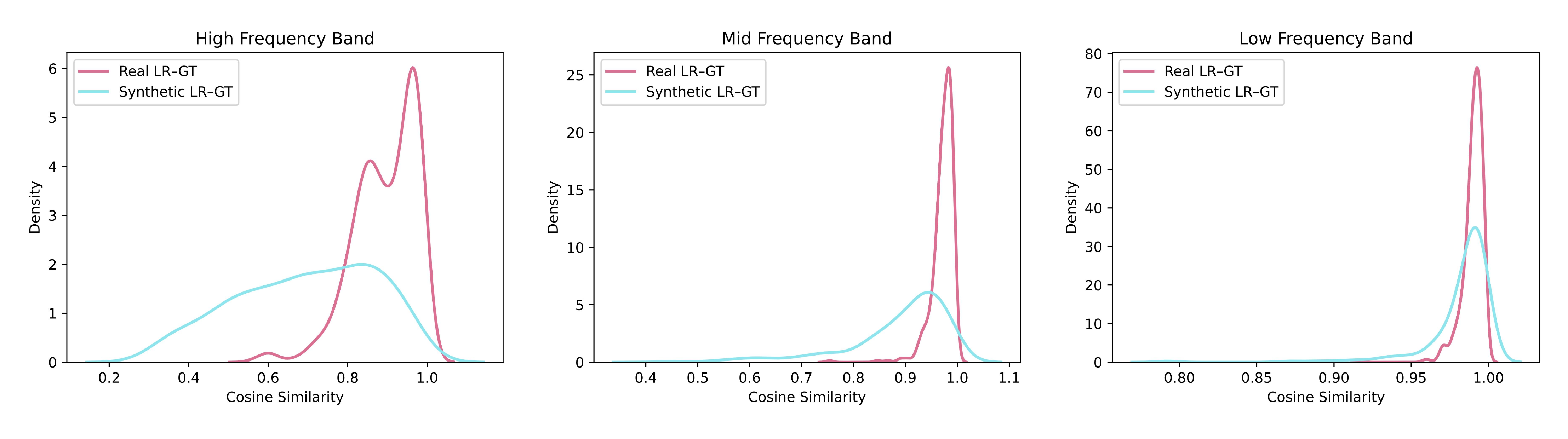}
  \vspace*{-1em}
  \caption{Kernel density estimation of cosine similarity distributions between LR–HR image pairs in deep feature space. Synthetic data preserves high similarity in low-frequency components but exhibits increasing distribution broadening and leftward shift in mid- and high-frequency bands, indicating greater domain shift in fine-grained details compared to real-world data.}
  \label{fig:add_analysis_DAE}
\end{figure*}

We further analyze the relationship between low-resolution and high-resolution (HR) image pairs in deep feature space by computing the cosine similarity between their feature embeddings extracted from a pretrained network. Kernel density estimation (KDE) is employed to estimate the distribution of cosine similarities across the dataset. 
Across different frequency bands, synthetic data exhibits a notable distribution expansion coupled with a significant domain shift, as shown in Figure~\ref{fig:add_analysis_DAE}. In the low-frequency band, synthetic LR–GT pairs maintain high cosine similarity concentrated near 1.0, closely matching real-world distributions and preserving global structures. However, as the frequency increases, the synthetic distribution becomes increasingly broader and shifts toward lower similarity values; specifically, the mid-frequency band reveals discrepancies in meso-scale textures, while the high-frequency band displays the most pronounced flattening and leftward shift. This trend suggests that while synthetic degradations cover a wider range of variations, they fail to accurately replicate the complex, fine-grained artifacts of real-world processes, identifying high-frequency components as the primary driver of the overall input distribution shift. Therefore, when performing reward feedback learning on real-world data, we primarily apply optimization at the final timestep. This is because, during early timesteps, the diffusion model focuses on global structural information, whereas later timesteps emphasize high-frequency details. Moreover, as verified in the Draft~\cite{draft}, optimizing at the final timestep also drives passive optimization of preceding timesteps.

\subsection{Training details}
We select two representative Diffusion-SR paradigms and conduct experiments across models of different scales: DiT4SR and ResShift. 
Specifically, DiT4SR is built upon Stable Diffusion 3.5 with 2716.8M parameters, which shares a similar architecture with Stable Diffusion 3~\cite{sd3}. The training process is conducted on 512 × 512 resolution images. We train our model with a constant learning rate of $1e^{-6}$ with a batch size of 8. During inference, we adopt the default sampling schedule of SD3.5 with 40 sampling steps (T). The scale of classifier-free guidance (CFG) is set to 8 in our experiments. Following~\cite{DiT4SR}, the prompt of the input LR image is obtained from LLaVA~\cite{liu2023llava}.  For DiT4SR, we update only the control network while keeping all other components frozen. For ResShift, we freeze the first $70\%$ of the network parameters to retain its pre-trained super-resolution capability, and only update the remaining $30\%$ during training.

ResShift is a lightweight model trained from scratch with 16.7M parameters, which shares a similar architecture with UNet~\cite{DDPM}, the training process is conducted on 256 × 256 resolution images.  We train our model with a constant learning rate of $1e^{-6}$ with a batch size of 8. During inference, we adopt the default sampling schedule of ResShift with 15 sampling steps (T).

All the evaluation metrics are implemented by PyIQA~\cite{pyiqa}. Note that the metric of ‘ClipIQA’ is implemented with the setting of ‘clipiqa+\_vitL14 512’ provided by PyIQA.

For the reward function $r$, we employ ClipiQA; for the distortion metric $D$, we adopt LPIPS; and for the preference loss function $\phi$, we use ReLU. For Semantic features function $f$, we use DINOv2 of 'dinov2\_vitl14\_reg4'. The parameter $\lambda_{\text{sem}}$ is set to $0.001$. We further set the reward weight to $0.0003$ before applying the dynamic weighting scheme $\lambda(t)$.

\section{Extended Ablation}
\subsection{Ablation on Loss Components}
To verify the effectiveness of our proposed structural constraint and semantic alignment, we conduct ablation studies on both DiT4SR and ResShift. For structural constraint, we employ perceptual supervision via LPIPS loss between the policy model's output and the ground truth (GT). We deliberately eschew conventional $L_1$ or $L_2$ constraints, as they tend to induce over-smoothed results. For semantic alignment, we leverage DINOv2 in the feature space to align the outputs of the reference and policy models. Since DINO captures high-level semantics, it effectively constrains structural distortions arising from reward hacking while reserving optimization overhead for pixel-level detail preferences. This dual-pronged approach ensures both the stability and efficacy of the training process.

As presented in Table~\ref{tab:app-ablation-loss-resshift}, utilizing solely the reward signal without additional constraints leads to a significant improvement in perceptual metrics. However, this gain is achieved at the expense of severe structural distortion, indicating that the model has succumbed to reward hacking. The introduction of either semantic or structural constraints results in a regression in perceptual scores but successfully mitigates distortion. Ultimately, by integrating both constraints, our model achieves a stable and synergistic improvement across both perceptual quality and distortion metrics.

\subsection{Ablation on Effectiveness of Bidirectional Reward-Guided Diffusion}
We perform the same ablation experiments on the smaller ResShift model as on DiT4SR in the main text, and find that its performance is highly sensitive to the configuration, as shown in Table~\ref{tab:app-ablation-bimix-resshift}.

\subsection{Dynamic Weight}
We also investigate the impact of different distortion–perception weighting strategies across diffusion timesteps on model performance on the smaller ResShift model, as shown in Table~\ref{tab:app-ablation-weight-resshift}.

\begin{table}[htbp]
    \footnotesize
    \renewcommand{\arraystretch}{1.1}
    \setlength{\tabcolsep}{5pt}
    \centering
    \caption{Ablation results on RealSR for ResShift. All
    variants are trained using the same settings as the full model. }
    \begin{tabular}{c|ccc|cc}
    \toprule
    Model & \begin{tabular}[c]{@{}c@{}} $\mathcal{L}_{\text{struct}}$ \end{tabular} 
          & \begin{tabular}[c]{@{}c@{}} $\mathcal{L}_{\text{sem-align}}$ \end{tabular} 
          & \begin{tabular}[c]{@{}c@{}} $\mathcal{L}_{\text{reward}}$ \end{tabular} 
    & ClipIQA $\uparrow$  & LPIPS $\downarrow$ \\ \midrule
    \rm{A}      & \XSolidBrush   & \XSolidBrush   &  \CheckmarkBold   & 0.535  & 0.399  \\
    \rm{B}      & \XSolidBrush   & \CheckmarkBold &  \CheckmarkBold   & 0.513  & 0.377  \\ 
    \rm{C}      & \CheckmarkBold & \XSolidBrush   &  \CheckmarkBold   & 0.472 &  0.327  \\ \midrule
     FULL       & \CheckmarkBold & \CheckmarkBold &  \CheckmarkBold   & 0.465 & 0.314    \\ \bottomrule
    \end{tabular}
    \label{tab:app-ablation-loss-resshift}
    \end{table}

\begin{table}[htbp]
    \footnotesize
    \renewcommand{\arraystretch}{1.1}
    \setlength{\tabcolsep}{15pt}
    \centering
    \caption{Ablation study on forward and backward optimization for ResShift on RealSR. We report training cost (GPU hours) and perceptual quality measured by ClipIQA and LPIPS. Note that GPU hours are normalized relative to 3 (set to 100\%). Due to the small model capacity, significant degradation in LPIPS often indicates that the model is likely to fall into reward hacking.}
    \begin{tabular}{c|cc|c}
    \toprule
    Setting & ClipIQA $\uparrow$ & LPIPS $\downarrow$ & \begin{tabular}[c]{@{}c@{}}Train cost\end{tabular} \\  \midrule
    1 & 0.462 & 0.315 & 23\% \\
    2 & 0.455 & 0.391 & 85\% \\
    3 & 0.492 & 0.352 & 100\% \\
    4 & 0.465 & 0.314 & 70\% \\ \bottomrule
    \end{tabular}
    \label{tab:app-ablation-bimix-resshift}
\end{table}

\begin{table}[htbp]
    \footnotesize
    \renewcommand{\arraystretch}{1.1}
    \setlength{\tabcolsep}{24pt}
    \centering
    \caption{Ablation study on different distortion–perception weighting for ResShift on RealSR. We report perceptual quality measured by ClipIQA and LPIPS.}
    \begin{tabular}{c|cc}
    \toprule
    Setting & ClipIQA $\uparrow$ & LPIPS $\downarrow$ \\  \midrule
    $\gamma=0.1$ & 0.477 & 0.331 \\
    $\gamma=0.5$ & 0.476 & 0.328 \\
    $\gamma=1.0$ & 0.475 & 0.324 \\
    $\gamma=2.0$ & 0.472 & 0.320 \\
    $\gamma=4.0$ & 0.468 & 0.316 \\
    $\gamma=8.0$ & 0.465 & 0.314 \\ \bottomrule
    \end{tabular}
    \label{tab:app-ablation-weight-resshift}
\end{table}

\section{More visual results}
\label{sec:app-More-vision}
\begin{figure*}[htbp]
      \begin{center}
    \includegraphics[width=1.0\linewidth]{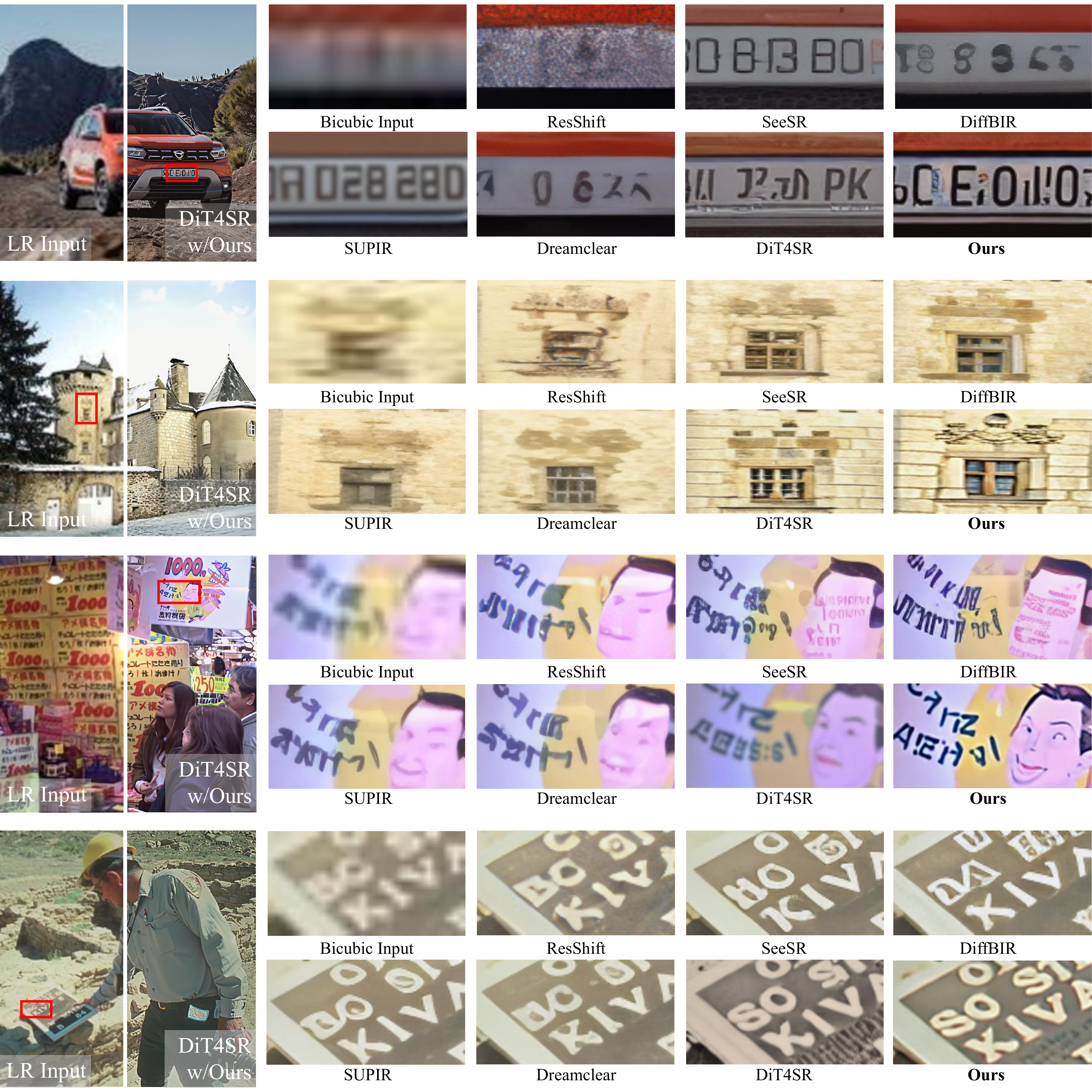}
      \end{center}
    \caption{Qualitative comparisons with state-of-the-art Real-ISR methods. Our method performs best in terms of image realism and detail generation especially preserving fine structures and restoring text details.}
    \label{fig:vision_v2}
    \vspace{-1em}
\end{figure*}

\begin{figure*}[htbp]
      \begin{center}
      \includegraphics[width=0.9\linewidth]{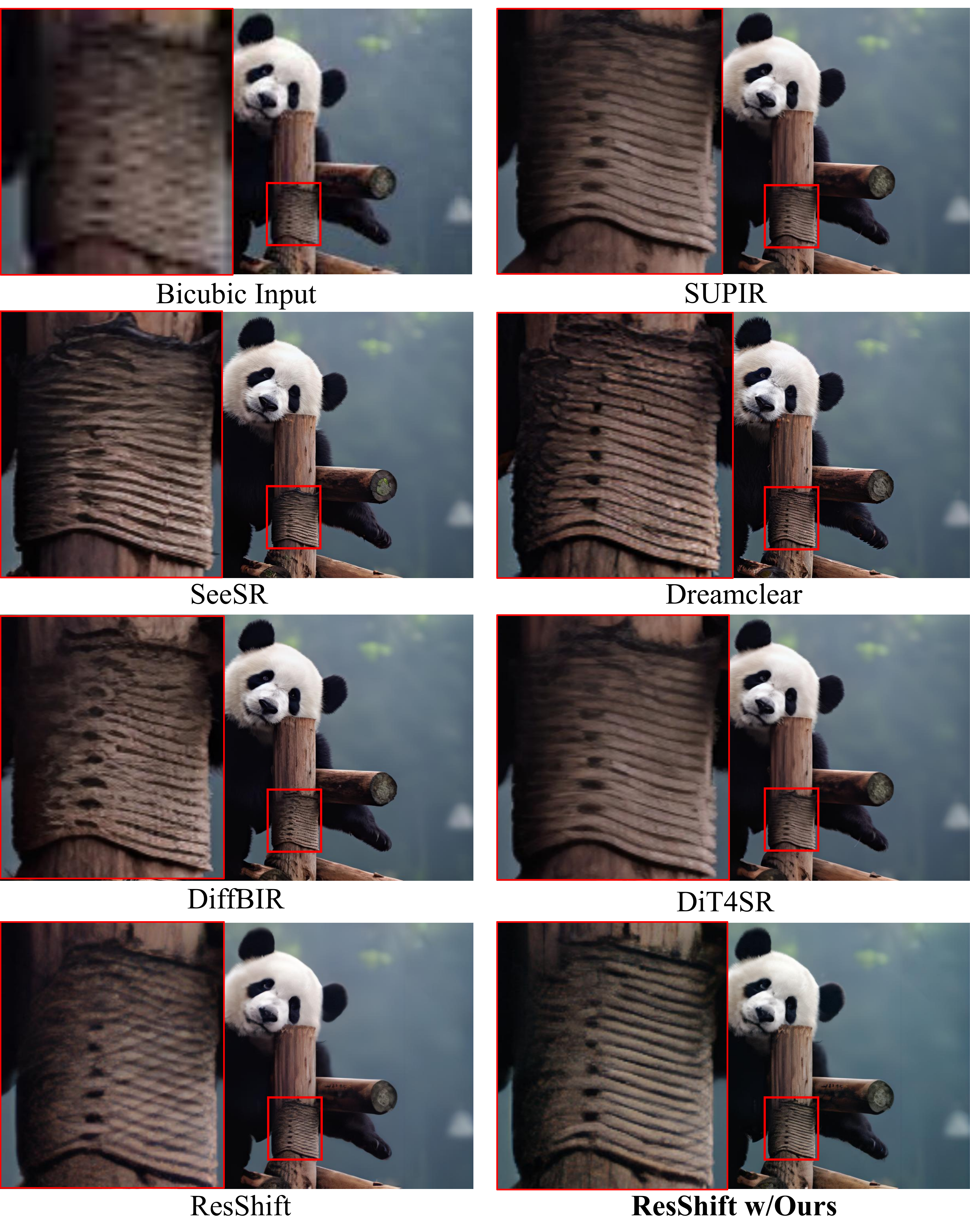}
      \end{center}
    \caption{Qualitative comparisons with state-of-the-art Real-ISR methods.}
    \label{fig:vision_v3}
\end{figure*}

\begin{figure*}[htbp]
      \begin{center}
      \includegraphics[width=0.9\linewidth]{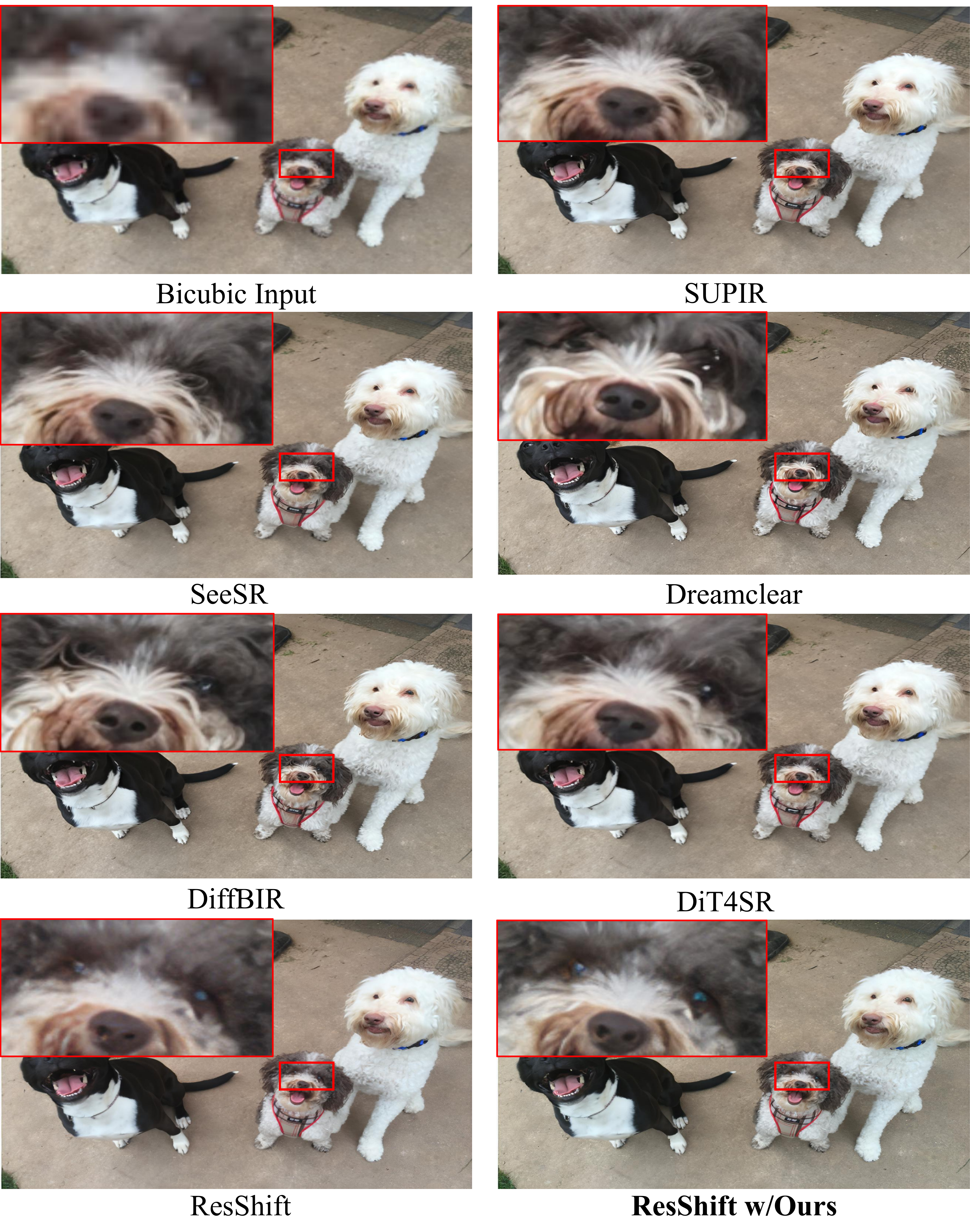}
      \end{center}
    \caption{Qualitative comparisons with state-of-the-art Real-ISR methods. }
    \label{fig:vision_v4}
\end{figure*}

\begin{figure*}[htbp]
      \begin{center}
      \includegraphics[width=0.9\linewidth]{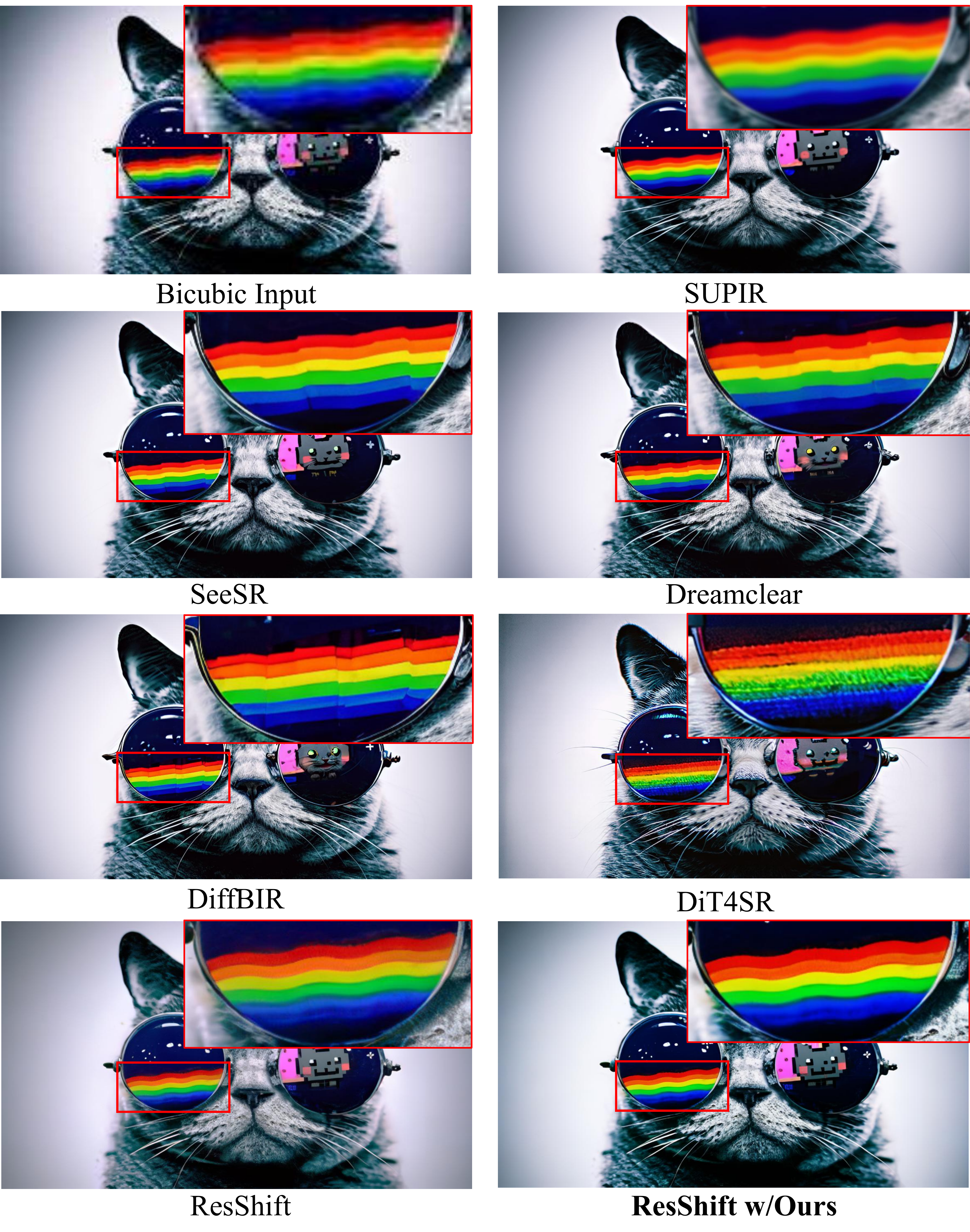}
      \end{center}
    \caption{Qualitative comparisons with state-of-the-art Real-ISR methods.}
    \label{fig:vision_v5}
\end{figure*}

\begin{figure*}[htbp]
      \begin{center}
      \includegraphics[width=0.9\linewidth]{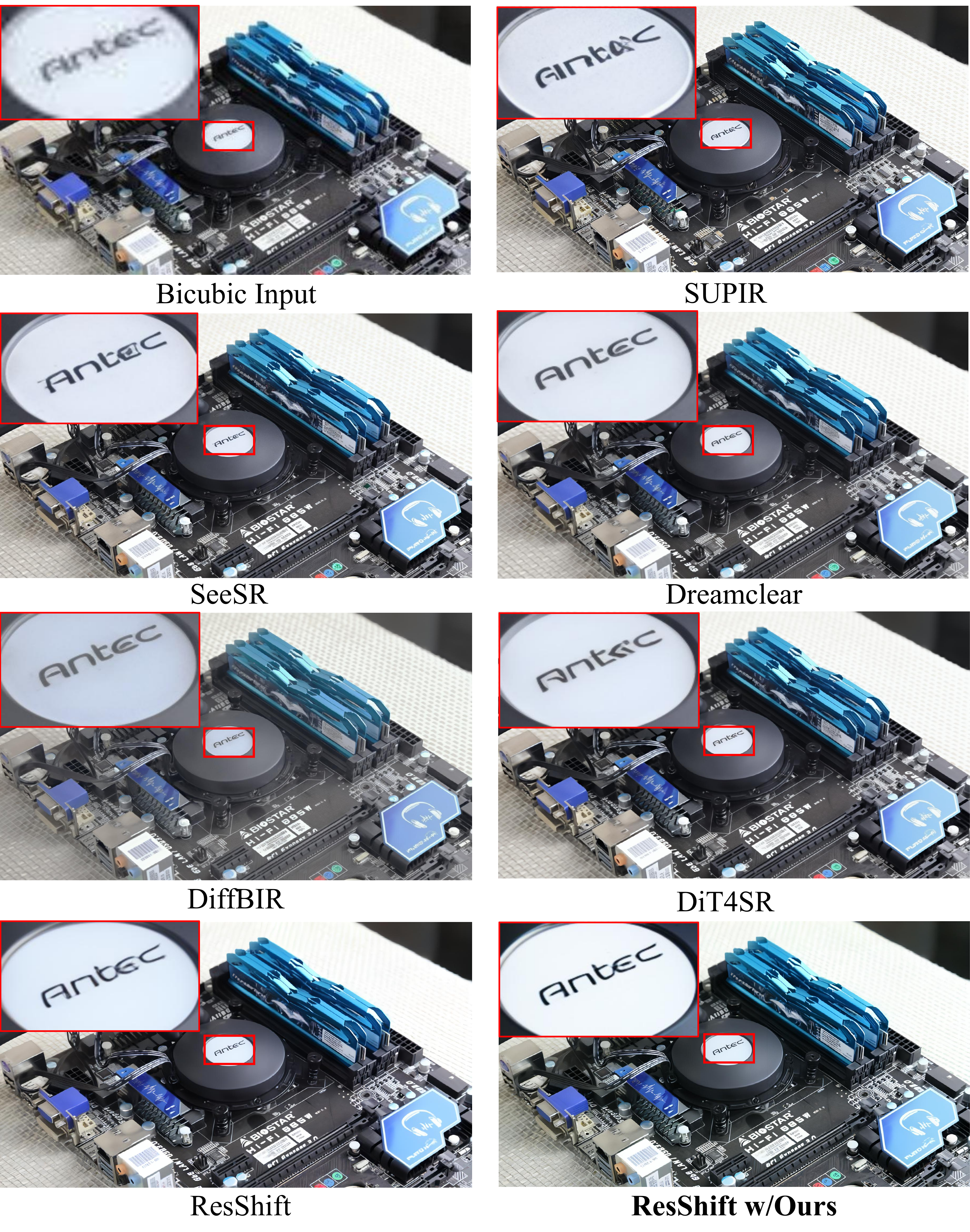}
      \end{center}
    \caption{Qualitative comparisons with state-of-the-art Real-ISR methods.}
    \label{fig:vision_v6}
\end{figure*}

We provide additional visual comparisons on real-world datasets to demonstrate the robustness of our method. As shown in Figure~\ref{fig:vision_v2}, Figure~\ref{fig:vision_v3}, Figure~\ref{fig:vision_v4}, Figure~\ref{fig:vision_v5} and Figure~\ref{fig:vision_v6}, our method generates more plausible details while effectively suppressing artifacts, yielding the best visual results.

\section{User Study Configuration}
User Study. To further validate the restoration quality, we invite 20 volunteers to conduct the user study. We randomly select 40 LR images from these four datasets (DrealSR, RealSR, RealLR200, and RealLQ250), and adopt the three latest methods (Baseline, SeeSR, DiffBIR, and DreamClear) for comparison.

In the user study, participants were presented with three images for each evaluation: the original LR input image, the restoration result generated by our method, and the restoration result from a randomly selected other method. 

\begin{figure*}[htbp]
  \centering
  \includegraphics[width=0.9\linewidth]{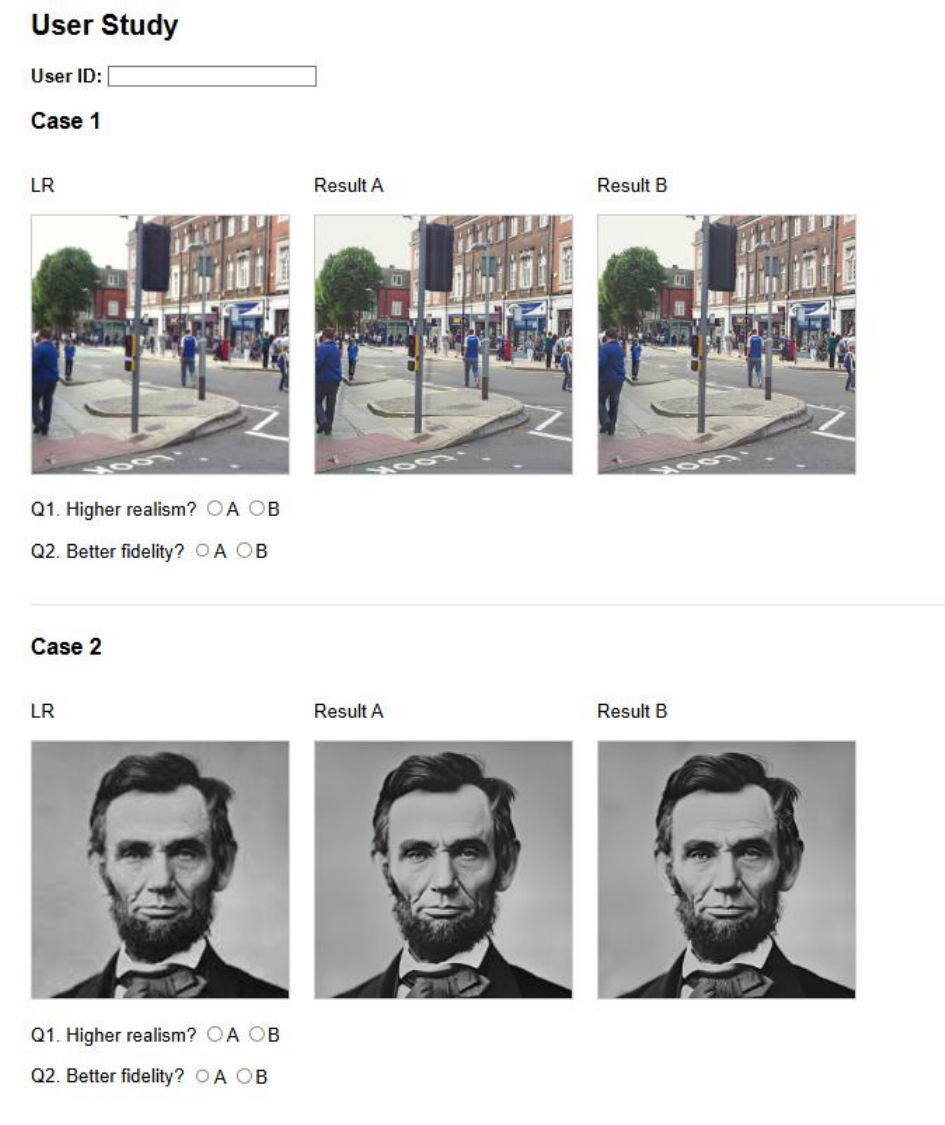}
  \caption{Comparison user study HTML example in the user study.}
  \label{fig:user_study}
\end{figure*}

Raters were asked to select the better image 
based on two distinct criteria: 
(1) Which restoration result has higher image realism? 
(2) Which restoration result has better fidelity to the original image content? 
The interface designed for our user study is illustrated in Figure~\ref{fig:user_study}.

\section{Related Methods}
Another line of research involves $D^2PO$~\cite{wu2025dposr}, which explores preference optimization via synthetic data rollouts. Although $D^2PO$ provides a robust baseline for aesthetic alignment, its application to Real-World ISR is limited by the neglect of LR-to-HR distribution gaps. Our Bird-SR framework treats this distribution shift as a primary constraint. Notably, our approach is orthogonal to $D^2PO$; while our Bird-SR provides the base preference learning objective, $D^2PO$'s HPO (Hierarchical Preference Optimization) and hybrid rewards can be seamlessly integrated into our Bird-SR framework to handle complex real-world degradations.

\section{Broader Impacts}
Bird-SR aims to improve image super-resolution by incorporating bidirectional reward feedback into diffusion-based training, enabling more effective adaptation to complex and diverse degradations. The proposed method has the potential to benefit a wide range of real-world applications that rely on high-quality visual information, including remote sensing, medical imaging, scientific visualization, and low-bandwidth image transmission, where enhanced resolution can facilitate downstream analysis and decision-making.

By explicitly modeling perceptual preferences through reward signals and leveraging both synthetic and real-world low-resolution data, Bird-SR may contribute to more robust and generalizable restoration models, reducing performance gaps between controlled benchmarks and practical deployment scenarios. This could lower the barrier for applying super-resolution techniques in resource-constrained or data-limited environments.

At the same time, as with other image restoration and enhancement methods, Bird-SR may introduce the risk of hallucinated details or perceptual biases, particularly when deployed in safety-critical domains such as medical or forensic imaging. To mitigate these risks, Bird-SR is designed to operate within the constraints of physically grounded diffusion processes, and its reward models can be tailored or restricted to domain-specific criteria to better align with task requirements. We emphasize that Bird-SR is intended to assist, rather than replace, human judgment in sensitive applications.

Finally, we note that Bird-SR does not inherently introduce new privacy or fairness concerns beyond those common to existing super-resolution approaches. Future work will explore more fine-grained and interpretable reward models, as well as mechanisms for uncertainty estimation and controllable generation, to further enhance the transparency and responsible deployment of reward-guided diffusion models.

\clearpage
{
    \small
    \bibliographystyle{ieeenat_fullname}
    \bibliography{main}
}


\end{document}